\UseRawInputEncoding
\documentclass{article}
\pdfoutput=1%For ARXIV
    \PassOptionsToPackage{numbers, compress}{natbib}
    \usepackage[final]{neurips_data_2024}

\usepackage[utf8]{inputenc} % allow utf-8 input
\usepackage[T1]{fontenc}    % use 8-bit T1 fonts
\usepackage{hyperref}       % hyperlinks
\usepackage{url}            % simple URL typesetting
\usepackage{booktabs}       % professional-quality tables
\usepackage{amsfonts}       % blackboard math symbols
\usepackage{nicefrac}       % compact symbols for 1/2, etc.
\usepackage{microtype}      % microtypography
\usepackage{xcolor}         % colors
\usepackage{graphicx}
\usepackage{subcaption}
\usepackage{multirow} 
\usepackage{wrapfig}
\usepackage{threeparttable}

\newtheorem{definition}{Definition}

\usepackage{hyperref}
\title{Job-SDF: A Multi-Granularity Dataset for Job Skill Demand Forecasting and Benchmarking}

\author{%
  Xi Chen$^{1}$\thanks{Equal contributions}, 
  Chuan Qin$^{2}$\footnotemark[1],
  Chuyu Fang$^{3}$,
  Chao Wang$^{1}$, \\
  \textbf{Chen Zhu}$^{1}$,
  \textbf{Fuzhen Zhuang}$^{4,5}$,
  \textbf{Hengshu Zhu}$^{2}$\thanks{Corresponding Authors},
  \textbf{Hui Xiong}$^{6,7\dagger}$\\
  $^1$University of Science and Technology of China \\
  $^2$Computer Network Information Center, Chinese Academy of Sciences \\
  $^3$Baidu Inc. $^4$Institute of Artificial Intelligence, Beihang University \\
  $^5$SKLSDE, School of Computer Science, Beihang University\\
  $^6$AI Thrust, The Hong Kong University of Science and Technology (Guangzhou) \\
  $^7$Department of Computer Science and Engineering,\\ The HongKong University of Science and Technology, Hong Kong SAR\\ 
  \texttt{chenxi0401@mail.ustc.edu.cn},\\
  \texttt{\{chuanqin0426, fangchuyu2022, chadwang2012, zc3930155\}@gmail.com},\\
  \texttt{zhuangfuzhen@buaa.edu.cn, zhuhengshu@gmail.com, xionghui@ust.hk} \\
}

\begin{document}

\maketitle

% \footnotetext[1]{Equal contributions}

\begin{abstract}

In a rapidly evolving job market, skill demand forecasting is crucial as it enables policymakers and businesses to anticipate and adapt to changes, ensuring that workforce skills align with market needs, thereby enhancing productivity and competitiveness. Additionally, by identifying emerging skill requirements, it directs individuals towards relevant training and education opportunities, promoting continuous self-learning and development. However, the absence of comprehensive datasets presents a significant challenge, impeding research and the advancement of this field. To bridge this gap, we present \textsl{Job-SDF}, a dataset designed to train and benchmark job-skill demand forecasting models. Based on millions of public job advertisements collected from online recruitment platforms, this dataset encompasses monthly recruitment demand.
Our dataset uniquely enables evaluating skill demand forecasting models at various granularities, including occupation, company, and regional levels. 
We benchmark a range of models on this dataset, evaluating their performance in standard scenarios, in predictions focused on lower value ranges, and in the presence of structural breaks, providing new insights for further research. Our code and dataset are publicly accessible via the \url{https://github.com/Job-SDF/benchmark}.

\end{abstract}
\section{Introduction}
Job skills encompass a range of abilities and competencies essential for performing tasks effectively in the workplace. These skills are broadly categorized into hard skills, such as technical and analytical abilities, and soft skills, including communication, teamwork, and adaptability~\cite{autor2010polarization}. Accurate forecasting of skill demand helps businesses and policymakers anticipate and address skill shortages and mismatches, and promotes skill development in high-demand areas, thereby supporting economic growth and stability~\cite{heckman2006effects,qin2023automatic}. By identifying emerging skill requirements, individuals are directed towards relevant training and education opportunities, fostering continuous self-learning and development to stay competitive in the labor market~\cite{kokkodis2021demand,zha2023career,zha2024towards,yu2024disco,sun2024large,dai2020enterprise,zhu2016days}. By tracking skill demand trends, employers gain deeper insight into recruits’ priorities, enhancing person-job fit.~\cite{qin2018enhancing,qin2019duerquiz,qin2020enhanced,yao2022knowledge,wu2024exploring,bian2020learning,luo2019resumegan,mashayekhi2024challenge,luo2018resumenet,shen2021joint,jiang2024enhancing}. Moreover, forecasting informs educational and training programs, ensuring that curricula align with the labor market's evolving needs~\cite{deming2017growing,chen2024collaboration,zhang2023relicd}.

Traditionally, skill demand analysis has relied on labor-intensive, survey-based methods limited to specific companies or occupations~\cite{survey::rasdorf2016data,skill_shortage::healy2015adjusting,skill_shortage::bellmann2014skill}. However, over the past decade, the rapid evolution of the internet has spurred the emergence of online recruitment platforms. These platforms have become the primary channels for job advertisements for numerous enterprises and organizations, accumulating vast amounts of job advertisement data. By leveraging this data, researchers have formulated skill demand forecasting as a time series task, utilizing various machine learning models such as autoregressive integrated moving average (ARIMA)~\cite{das2020learning}, recurrent neural networks (RNNs)~\cite{garcia2022practical}, and dynamic graph autoencoders (DyGAEs)~\cite{dl::skill-demand-supply}, to predict future skill needs. 
%Over the past decade, the rapid development of the internet has led to the rise of online recruitment platforms, which have become the primary channels for job advertisements for many enterprises and organizations. This shift has resulted in the accumulation of vast amounts of job advertisement data. 
%Recently, studies have employed this data and integrated machine learning techniques, such as Support Vector Machine (SVM) and XGBoost, for predicting skill shortages and measuring skill demand~\cite{skill_shortage::dawson2020predicting, r::hirvonen2022new,r::vankevich2020ensuring}.
%To further forecast future skill demand, some research has utilized deep learning methods, including Recurrent Neural Network (RNN)-based models and Dynamic Graph Autoencoder (DyGAE), to capture the dynamic nature of skill demand, thus demonstrating efficacy~\cite{dl::chao2024cross,dl::skill-demand-supply}.

A major challenge impeding progress in this field is the lack of comprehensive and publicly accessible datasets. Existing studies do not provide open-source datasets, making it difficult for researchers to replicate experimental results and identify bottlenecks in current research. Furthermore, these datasets primarily focus on predicting skill demand variations across different occupations, with a notable lack of modeling and prediction at other granularities, such as companies or regions. This limitation hinders comprehensive comparisons between different models and impedes the exploration of potential downstream applications, such as human capital strategy development and regional policy formulation. Additionally, the significant variations in skill demand present further challenges. Existing studies, which rely on metrics such as Mean Squared Error (MSE), struggle to evaluate the performance of skill demand forecasting models for low-frequency skill terms. For instance, some emerging skills, such as large language models (LLMs), may initially show low demand but are crucial for the job market due to their potential to reshape existing occupations.

To this end, in this paper, we propose \textsl{Job-SDF}, a multi-granularity dataset designed for job skill demand forecasting research. Specifically, we collected millions of public job advertisements from online recruitment platforms. By extracting skill terms from job advertisement texts, we quantified the monthly skill demand at various granularities, including occupations, companies, and regions, to construct our dataset. This dataset encompasses 2,324 types of skills, 52 occupations, 521 companies, and 7 regions. We then use the Job-SDF dataset to benchmark a wide range of models for job skill demand forecasting tasks at various granularities.
%time-series forecasting models both in job skill demand forecasting and zero-shot job skill demand forecasting tasks, at various granularities. 
These models include statistical time series models (e.g., ARIMA~\cite{arima:ariyo2014stock}), deep learning-based methods such as RNN-based models~\cite{lin2023segrnn,dl::skill-demand-supply}, Transformer-based models~\cite{zhou2021informer,wu2021autoformer,liu2022non,zhou2022fedformer,kitaev2020reformer}, MLP-based models~\cite{zeng2023transformers,wang2024timemixer}, as well as several state-of-the-art time-series forecasters~\cite{zhou2022film,liu2024koopa}. Performance is evaluated using regression metrics such as Mean Absolute Error (MAE) and Root Mean Squared Error (RMSE). Additionally, we use the Symmetric Mean Absolute Percentage Error (SMAPE)~\cite{sareminia2023support} and Relative Root Mean Squared Error (RRMSE)~\cite{chen2017new} metrics to account for the significantly varying nature of skill demand values, which is particularly useful for evaluating model performance in predicting lower value ranges. Moreover, we further investigate the impact of structural breaks in job skill demand time series data on model performance. The Job-SDF dataset, along with data loaders, example codes for different models, and evaluation setup, are publicly available in our GitHub repository: \url{https://github.com/Job-SDF/benchmark}. %To our knowledge, Job-SDF is the first publicly available dataset and benchmarking platform specifically for job skill demand forecasting.

\section{Related Work}
%  related work写法参考：Federated Rule Dataset for Rule Recommendation Benchmarking：看一下intro，先介绍SDA分析的价值，以往是什么。最近由于平台数据、很多基于ML的方法，例如。。、其中，一些研究利用NN/DL，取得了。。。 例如，。。。。。 存在问题是什么。。。

Skill demand forecasting can analyze how skills evolve over time, aiding experts in evaluating technological advancements~\cite{manning2004we, abowd2007technology,qin2023comprehensive}, assessing wage inequality~\cite{juhn1999wage,lee2015technological,sun2021market}, and generating employment opportunities~\cite{benson2011raising}. Furthermore, the skills required in the 21st-century workplace will differ significantly from those in previous eras~\cite{hilton2008research}. Predicting skill demands benefits personal career transitions and corporate management strategies.

% \textcolor{red}{Traditional research on skill demand forecasting relies on survey data, which is labor-intensive and limited to specific occupations or companies~\cite{survey::rasdorf2016data, skill_shortage::healy2015adjusting, skill_shortage::bellmann2014skill}. With the development of online recruitment platforms, some data-driven methods using collected job advertisements have employed machine learning techniques to forecast skill demand. For example, Support Vector Machines (SVM) leverage text information to measure skill demand~\cite{r::hirvonen2022new}. Accounting for temporal patterns, XGBoost is used to predict future occupational skill shortages~\cite{skill_shortage::dawson2020predicting}.

Recently, with the rapid accumulation of data and continuous advancements in technology, skill demand forecasting has demonstrated significant vitality. 
\emph{Das et al.} proposed a method for dynamic task allocation to investigate the evolution of job task requirements over a decade of AI innovation across different salary levels ~\cite{das2020learning}.
Given the effectiveness of RNN in multi-step prediction, some researchers have integrated skill demand forecasting with RNN algorithms, achieving promising results ~\cite{garcia2022practical, lin2023segrnn}. 
In addition, considering the supply-demand dynamics of the labor market concurrently, CHGH designed a joint prediction model based on the encoder-decoder architecture to achieve trend prediction for both skill supply and demand sides~\cite{dl::skill-demand-supply}.
Moreover, to capture the dynamic information of occupations, a pretraining-enhanced dynamic graph autoencoder has been developed to efficiently forecast skill demand at the occupational granularity~\cite{chen2024pre}. 
% However, these research efforts have focused on predicting skill demand in the macro market or at a certain granularity. The lack of unified data is not conducive to benchmarking across different granularities. Moreover, the absence of publicly available research data hinders the reproduction of experimental results. This makes it difficult to compare and improve current models, which significantly impedes the development of this research field.}

However, the predominance of closed-source datasets has significantly elevated the barrier of researchers and constrained the pace of methodological advancements. While open-source skill-related datasets such as O*NET~\cite{cifuentes2010use} and ESCO~\cite{cosgrove2024mapping} provide skill taxonomies, they do not quantify skill demand. Furthermore, the current research data focuses either on macro-market skill demand predictions or analyses at a specific granularity, neglecting multi-level labor market analysis. This limitation generally hampers the transferability of the modeling approaches.

\section{Job-SDF Dataset}
% 参考SubseasonalClimateUSA，要注意license那些。
\label{section3}
The Job-SDF dataset is built from job advertisements collected on online recruitment platforms, encompassing dynamic job skill demand time series data at various granularities, recorded monthly. The dataset is \href{https://creativecommons.org/licenses/by-nc-sa/4.0/deed.en}{CC BY-NC-SA~4.0} licensed, accessible via the URL \url{https://github.com/Job-SDF/benchmark}. We summarize the dataset construction process, task description, and dataset analysis below.
%The Job-SDF dataset collects online recruitment advertisement data of leading companies and forms dynamic skill demand sequences at various granularities on a monthly scale. The dataset is regularly updated, CC BY 4.0 licensed. We summarize dataset construction steps below and provide supplementary details in the appendix.

\subsection{Data Collection and Processing}
\setcounter{footnote}{0}
\noindent\textbf{Job Advertisement Collection.}
We collected public job advertisements for 52 occupations from 521 companies on online recruitment platforms. We obtained unique records after removing identical job advertisements posted simultaneously by different companies on various platforms. Each record contains five types of information: (1) \textit{Job Requirement}, which is a text segment that outlines the specific skills required of candidates applying for the job; (2) \textit{Company}, which identifies the company that posted the job advertisement; (3) \textit{Occupation}, which specifies the job advertisement's category. Our dataset encompasses 52 detailed occupations (L2-level), such as front-end development engineer and financial investment analyst. Additionally, these 52 occupations are grouped into 14 broader categories (L1-level); (4) \textit{Region}, which indicates the primary geographic divisions in China where the job postings are located. These regions are classified based on their geographical orientation;
% and include Central China, South China, North China, East China, Northeast China, Southwest China, and Northwest China; 
(5) \textit{Posting Time}, which records the date when the job was posted, including the year, month, and day.
%We initially selected 521 companies and collected all job advertisements they posted publicly online from January 2021 to December 2023. 
%After accounting for duplicate postings by these companies, we filtered out redundant entries and obtained 10.35 million unique job advertisements as the job posting data. The data includes five fields: \textit{Job Description}, \textit{Company}, \textit{Occupation}, \textit{Region} and \textit{Posting Time}. 
% The definitions of these fields are as follows:
% (1) \textit{Job Description} includes the job requirements, with skill requirements typically extractable from this text information.
% (2) \textit{Company} identifies the company that posted the job advertisement.
% (3) \textit{Occupation} specifies the major occupation category to which the job belongs. Our data encompasses 52 major occupations, such as front-end development engineer and financial investment.
% (4) \textit{Region} indicates the primary geographic divisions in China where the job postings are located. These regions are classified based on their geographical orientation and include Central China, South China, North China, East China, Northeast China, Southwest China, and Northwest China.
% (5) \textit{Posting Time} records the date when the job was posted, including the year, month, and day.

% extracts skill demand in each job posting, quantifies the skill demand across various granularities, and outputs a standardized collection of machine-learning-ready Python Pandas DataFrames.

\textbf{Job Skill Extraction.} 
%After obtaining the job advertisement data, we explicitly extracted the skill requirements from the \textit{Job Requirement} in each job advertisement using a Named Entity Recognition (NER) model~\cite{fang2023recruitpro}.
After acquiring the job advertisement data, we utilized a Named Entity Recognition (NER) model, as referenced in~\cite{qin2022towards,fang2023recruitpro,yao2021interactive,jiang2024towards}, to explicitly extract skill requirements from the \textit{Job Requirement} of each advertisement. 
Specifically, we first annotated a dataset for training the NER model by identifying skill terms within the job requirement texts. To achieve this, we devised a set of regular expressions tailored to the characteristics of skill descriptions and used these to match skill words in job advertisements. Subsequently, we merged all matched skill words to formulate a raw skill dictionary, including their corresponding frequencies across job advertisements. We then filtered out low-frequency words and manually annotated the raw skill dictionary to create a refined skill dictionary. Along this line, we excluded unreasonable skill words matched by the regular expressions that did not appear in the refined skill dictionary, establishing an initial correspondence between the \textit{Job Requirement} and the skill requirements. 
%First, we annotated a dataset by mapping the \textit{Job Requirement} to skill requirements for training the NER model. 
%To achieve this, we devised a comprehensive set of regular expressions tailored to the characteristics of skill descriptions and used these to match skill words in job advertisements. Subsequently, we merged all matched skill words to formulate a raw skill dictionary, including their corresponding frequencies across job advertisements. We then filtered out low-frequency words and manually annotated the raw skill dictionary to create a refined skill dictionary. Furthermore, we excluded unreasonable skill words matched by the regular expressions that did not appear in the refined skill dictionary, establishing an initial correspondence between the \textit{Job Requirement} and skill requirements.

Based on this annotated data, we trained an NER model to extract required skills from the \textit{Job Requirement} section for all job advertisements. Experts then aggregated the skills extracted by the NER model based on their meaning and content, grouping together those with similar meanings or repeated expressions. This process resulted in a skill dictionary $\mathcal{S}$ of 2,324 standardized skill words, mapping original skill word descriptions to standardized skill words. The skill dictionary was then used to filter and map the skill words extracted by the NER model, ultimately obtaining standardized skill requirements for each job requirement. These standardized requirements were added to the job advertisement data as a new field, \textit{Skill Requirements}.

%Based on this annotated data, we trained a NER model to extract comprehensive required skills from the \textit{Job Requirement}. Using the skills extracted by the NER model, experts aggregated them based on their meaning and content, grouping together those with similar meanings or repeated expressions. This process resulted in a skill dictionary of 2,324 standardized skill words, mapping original skill word descriptions to standardized skill words. The skill dictionary was then used to filter and map the skill words extracted by the NER model, ultimately obtaining standardized skill requirements for each job description. These standardized requirements were added to the job advertisement data as a new field, \textit{Skill Requirements}.

\textbf{Job Skill Demand Estimation.}
Generally, the demand for different skills in the job market can be estimated by the volume of job advertisements listing these specific skills as requirements within a given time period~\cite{dl::skill-demand-supply}. Formally, given job advertisement data $\mathcal{P}=\{\mathcal{P}_1, ..., \mathcal{P}_t, ..., \mathcal{P}_T\}$, where each $\mathcal{P}_t$ represents the job advertisements posted at timestamp $t$, we use $D_{s,t}=\sum_{p \in \mathcal{P}_t} \mathbf{1}(s \in p)$ to estimate the demand for skill $s \in \mathcal{S}$ at time $t$. $s \in p$ indicates that job advertisement $p$ requires skill~$s$. 

Along this line, we can calculate skill demand at various granularities, such as occupation and company levels. We define the sets of L1-level occupations, L2-level occupations, companies, and regions as $\mathcal{A}^{o_1}$, $\mathcal{A}^{o_2}$, $\mathcal{A}^{c}$, and $\mathcal{A}^{r}$, respectively. The demands for skill $s$ at time $t$ under granularity $i \in \{o_1, o_2, c, r\}$ is then defined as follows:
\begin{equation}
D_{s, t}^i = [D_{s, t, a^i}^i]_{a^i \in \mathcal{A}^i}, \ \ D_{s, t, a^i}^i = \sum_{p \in \mathcal{P}_t} \mathbf{1}(s \in p) \cdot \mathbf{1}(a^i \in p),
\end{equation}
where $a^i \in p$ represents a job advertisement $p$ containing the attribute $a^i$ under granularity $i$. Similarly, we can further define skill demands $D_{s, t}^{i,j,...,k}$ across multiple granularities $\{i, j, ..., k\}$ by calculating:
\begin{equation}
D_{s, t, \overline{a}}^{i,j,...,k} = \sum_{p \in \mathcal{P}_t} \mathbf{1}(s \in p) \cdot \mathbf{1}(a^i \in p \land a^j \in p \land ... \land a^k \in p),
\end{equation}
where $\overline{a} = \{a^i, a^j, ..., a^k\}$, $a^i \in \mathcal{A}^i, a^j \in \mathcal{A}^j, ..., a^k \in \mathcal{A}^k$, and $D^{i,j,\ldots,k}_{s,t} \in \mathbb{R}^{ |\mathcal{A}^i||\mathcal{A}^j|\ldots|\mathcal{A}^k|}$.

\subsection{Job Skill Demand Forecasting Tasks}
We study model performance through job skill demand forecasting tasks at different granularities, including single and multiple levels. The primary goal of these tasks is to predict future job skill demands based on historical time series data of various skills. Formally, we have:

%To maximize the utility of our dataset for predicting future skill demand, we first conceptualize a general job skill demand forecasting task. Subsequently, we categorize the skills into two distinct groups based on their historical frequency and formulate the few-shot and zero-shot job skill demand forecasting tasks.

% \paragraph{Job Skill Demand Forecasting}
% Job skill demand forecasting aims to predict the demand for various skills across multiple granularities. 

\begin{definition}[Job Skill Demand Forecasting]
Given a granularity or a set of granularities $g$ and the observed job skill demand series from the previous $K$ timestamps, i.e., $\{D^{g}_{:,t-K+1}, ...,  D^{g}_{:,t}\}$, the goal of job skill demand forecasting is to learn a forecasting model $\mathcal{M}$ to predict the demand values for the next $H$ timestamps, denoted by $\{\hat{D}^{g}_{:,t+1}, \ldots, \hat{D}^{g}_{:,t+H}\}$.
% , that is,
% \begin{equation}
% \hat{D}^{g}_{:,t+1}, \ldots, \hat{D}^{g}_{:,t+H} = \mathcal{M}(D^{g}_{:,t-K+1},\ldots,D^{g}_{:,t-1};\Phi).
% \end{equation}
% Given a set of granularities $g = (i,j,\ldots,k)$ and the observed demand series of previous $K$ timestamps $D^{g}_{:,t-K+1},\ldots,D^{g}_{:,t-1}$, the task of job skill demand forecasting aims to predict the demand values for the next $H$ timestamps, denoted by $\hat{D}^{g}_{:,t+1}, \ldots, \hat{D}^{g}_{:,t+H}$. Theses values can be inferred by the forecasting model $\mathcal{F}$ with parameters $\Phi$ and a skill co-occurrence graph $\mathcal{G}^{g}$ as prior.
% \begin{equation}
% \hat{D}^{g}_{:,t+1}, \ldots, \hat{D}^{g}_{:,t+H} = \mathcal{F}(D^{g}_{:,t-K+1},\ldots,D^{g}_{:,t-1};\mathcal{G}^{g};\Phi)
% \end{equation}
\end{definition}
Our dataset includes skill demand time series data for L1-level occupations, L2-level occupations, companies, regions, and their combinations. We follow a standard protocol~\cite{MTS::wang2022learning} that categorizes all time-series data into training, validation, and test sets in chronological order with a ratio of 9:1:2. In the main text, we demonstrate results with $K$ set to 6 months and consider $H$ as 3 months to evaluate the performance of different forecasting models. More settings and results can be found in the Appendix \ref{app::add_exp} and our project repository. Based on the Job-SDF dataset, other researchers can easily adjust the parameters to suit their research objectives.

\begin{figure}
    \centering
    \begin{subfigure}{1\textwidth}
        \centering
        \includegraphics[width=\textwidth]{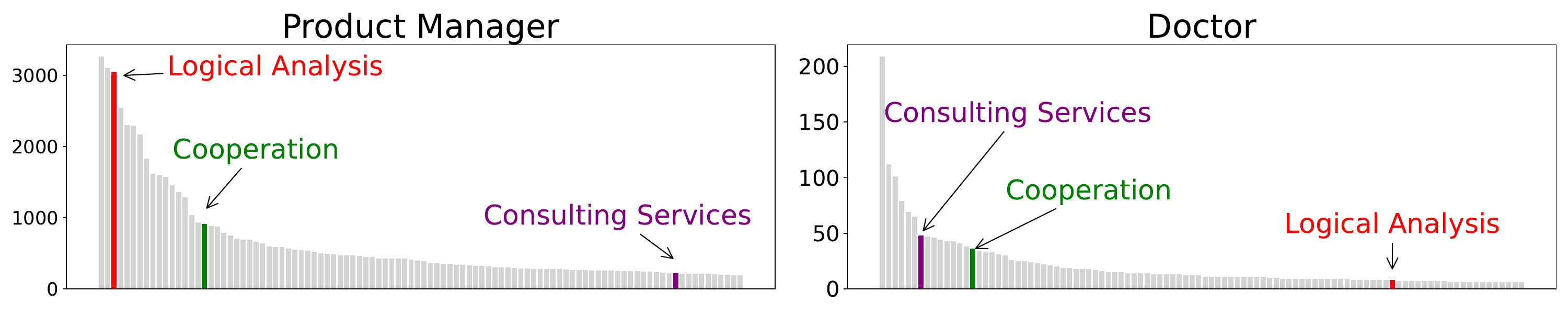}
        \caption{The skill demands under two occupations.}
        \label{fig_longtail}
    \end{subfigure}
    \vfill
    \begin{subfigure}{1\textwidth}
        \centering
        \includegraphics[width=\textwidth]{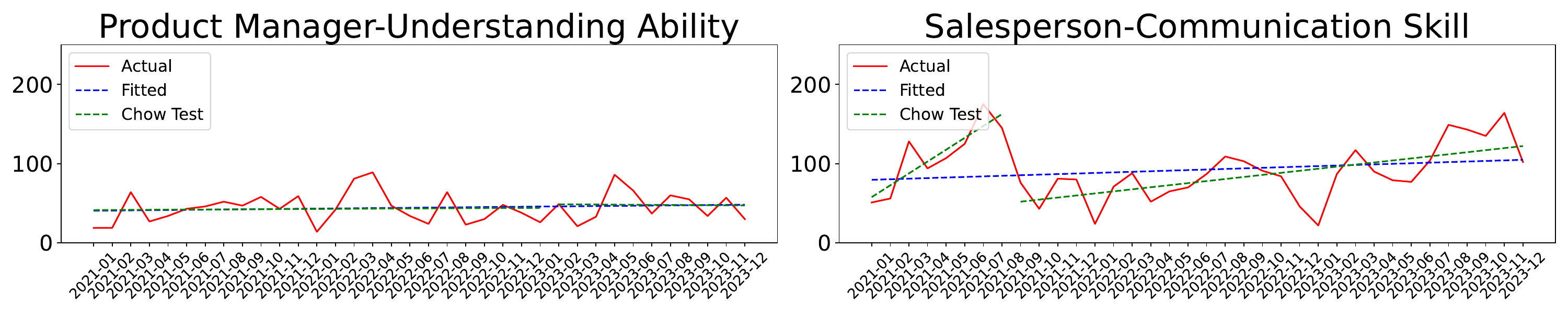}
        \caption{Appying the Chow test to two skill demand time series.}
        \label{fig_emergency}
    \end{subfigure}
    \caption{Data analysis on \textsl{Job-SDF}. (a) illustrates the long-tail phenomenon of skill demands under the product manager and doctor occupations. (b) illustrates the results under the Chow test for the absence (left) and presence (right) of structural breaks.}
    \vspace{-5mm}
    \label{fig:three_images}
\end{figure}
\subsection{Dataset Analysis}

\textbf{Varying Nature of Skill Demand.}
The values of skill demand exhibit significant differences and generally follow a long-tail distribution. This indicates that, at a specific granularity, only a few skills have high demand, while a wide range of skills are required by a limited number of jobs. For instance, Figure~\ref{fig_longtail} presents the skill demands under the product manager and doctor occupations in December 2022. The results clearly demonstrate the varying nature of skill demand values. This suggests that relying solely on metrics like RMSE to evaluate forecasting models' performance may overlook the prediction accuracy for low-frequency skills.

%Job skill demand follows a long-tailed distribution at each timestamp, indicating that, at a specific granularity, only a few skills maintain high demand, while a broad array of skills are needed by only a limited number of jobs. To investigate this long-tail phenomenon in the Job-SDF dataset, we compare the distribution of skill demand across two different occupations, as depicted in Figure \ref{fig:data_analysis}. The results clearly illustrate the presence of a long-tail distribution, where the head represents popular skills, and the tail comprises less popular, yet equally important skills. For instance, while the demand for a particular skill, such as xxx, may be low in occupation xxx, it is widely used in xxx and holds the potential for increased future demand in xxx.
%The long-tail distribution complicates the effective capture of demand for less popular skills, as a few highly demanded skills dominate the distribution. For example, employing common loss functions such as Mean Squared Error (MSE), which emphasizes larger errors, causes the model to struggle with fitting long-tailed samples accurately.

%call back zero-shot
\textbf{Structural Break Phenomenon.} As the labor market evolves, job skills that are not widely required today may become crucial in the future, while those currently in high demand may be supplanted by others. This dynamic can induce significant changes in the statistical properties of skill demand time series at various points in time. These changes may be reflected in the mean, variance, trend, or autocorrelation structure of the series. This phenomenon is known as structural breaks. A common method for detecting structural breaks is the Chow test, which evaluates whether there are significant differences in the regression coefficients across different periods~\cite{chow1960tests}. Figure~\ref{fig_emergency} illustrates the application of the Chow test in detecting structural breaks in various skill demand time series. The presence of structural breaks can impact the predictive accuracy of forecasting models. Further discussion will be provided in the experimental section.

\begin{wrapfigure}[16]{r}{0.43\textwidth}
    \begin{minipage}[t]{1\linewidth} 
    \centering
    \vspace{-5mm}
    \includegraphics[width=\textwidth]{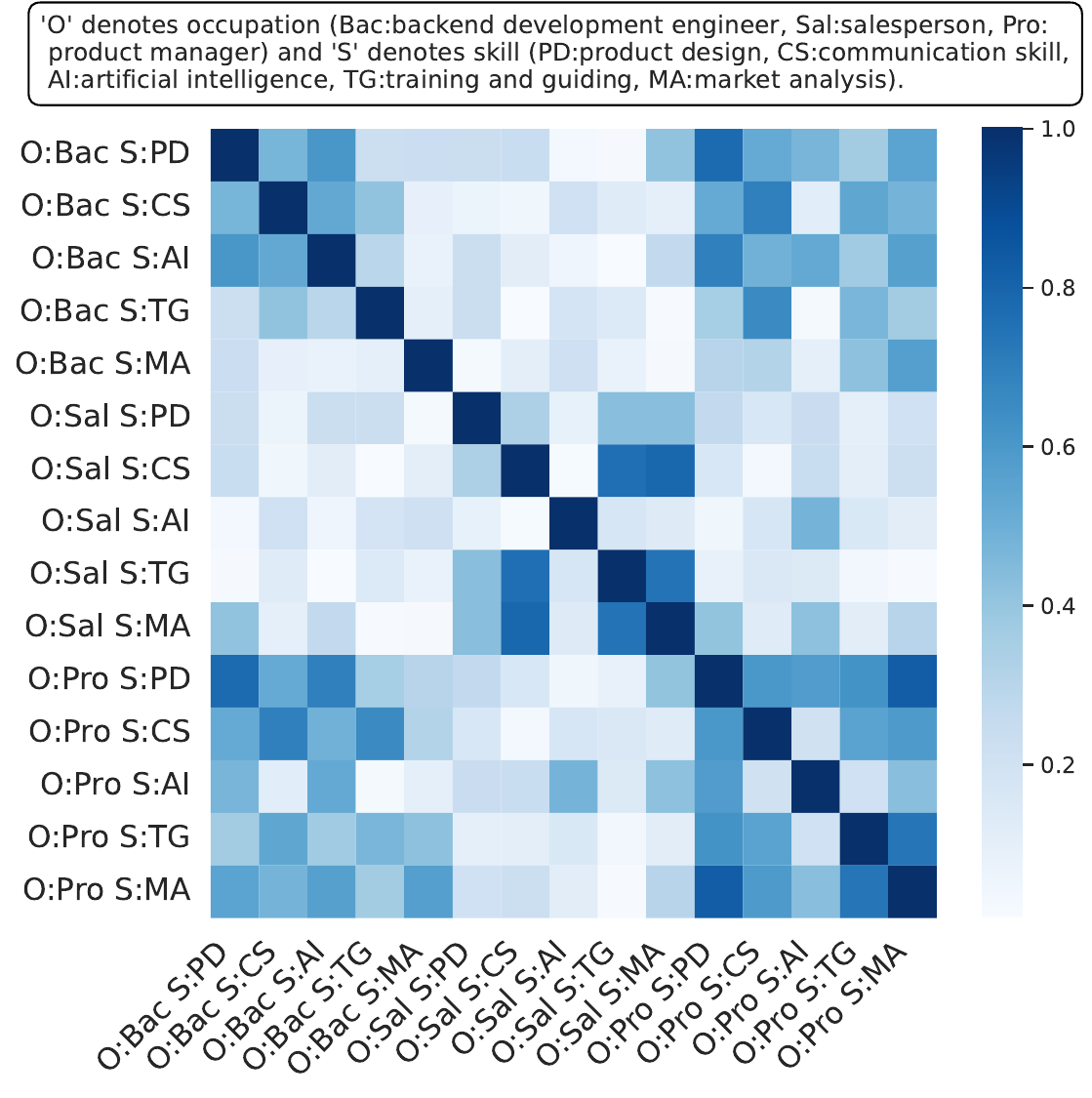}
    \vspace{-5pt}
    \vspace{-5mm}
    \caption{Pearson Correlation Coefficients.}
    \label{fig_heatmap}
    \end{minipage}
\end{wrapfigure}

\textbf{Inter-Series Correlation.}
Intuitively, the proposed job skill demand forecasting tasks can be categorized as multivariate time-series forecasting tasks~\cite{cao2020spectral}. Figure~\ref{fig_heatmap} shows the absolute values of the Pearson correlation coefficients of different skill demand series for the backend development engineer, salesperson, and product manager occupations. We found that the time series data for some skills exhibit significant correlation within the same occupation (i.e., product design and market analysis in product manager), as well as for the same skills across different occupations (i.e., product design in backend development engineer and product manager). This demonstrates the necessity of considering all variables as inputs for job skill demand forecasting models, as it captures the interrelationships among variables, preventing the loss of critical information when variables are considered in isolation.

% Rapidly evolving technologies give rise to emerging skills~\cite{challenge::garcia2022practical, challenge::yazdanian2022radar,walter2024european}. Skills that are currently underutilized or not in demand may become essential in the future. Since the demand for these skills has historically been minimal, accurately predicting which skills will see a surge in demand is highly challenging.
% To verify the widespread nature of this phenomenon, we analyzed two proportion curves shown in Figure \ref{fig:data_analysis}. The first curve represents the proportion of skills that had never been in demand in the year prior to each time slice, and the second curve shows the proportion of these skills that will be in demand in the next six months. Both curves maintain a consistent proportion, indicating that many skills rarely in demand before will be widely in demand in the future.
% In the following section, we additionally design zero-shot scenarios specifically for predicting the demand for these emerging skills.

    % \begin{subfigure}{0.33\textwidth}
    %     \centering
    %     \includegraphics[width=\linewidth]{figs/heatmap.pdf}
    %     \caption{Heatmap of Pearson Correlation Coefficients for Different Skills}
    %     \label{fig:image1}
    % \end{subfigure}

\textbf{Dataset Limitation.}
Recent studies~\cite{guo2023learning} suggest that incorporating relationships between different variants can enhance the performance of multivariate time-series forecasting. However, due to the lack of prior knowledge, our Job-SDF dataset does not yet include a graph of relationships between skills, such as predecessor-successor relationships. Instead, we constructed a skill graph based on the co-occurrence of skills in job advertisements from the training data. This graph is included in our dataset.

\section{Benchmark}
\subsection{Benchmark Models}
We evaluated several SOTA time-series learning models using our proposed Job-SDF dataset. These models are categorized into six groups based on their underlying architectures: statistical time series models, RNN-based models, Transformer-based models, MLP-based models, Graph-based models, and Fourier-based models. The implementation details for each model are provided in the Appendix \ref{app::add_exp}, and the open-source model implementations are available on \href{https://github.com/Job-SDF/benchmark}{our GitHub repository}.

%We assess a selection of prominent baseline models for time-series learning on our proposed Job-SDF dataset. These models are categorized into six groups according to their underlying architectures, encompassing statistical time series models, RNN-based architectures, Transformer-based approaches, MLP-based frameworks, Graph-based methodologies, and Fourier-based techniques.

\textbf{Statistical Time Series Model.} We first consider two statistical methodologies, namely \textit{ARIMA}~\cite{box2015time} and \textit{Prophet}~\cite{taylor2018forecasting}, both of which have been widely used in various contexts. The ARIMA model, which integrates differencing and moving averages within autoregression, has proven effective in forecasting occupational task demands~\cite{das2020learning}. \textit{Prophet} decomposes time series data into trend, seasonality, and holiday components, allowing it to handle both linear and nonlinear trends with changepoints. However, these models often struggle to capture complex nonlinear relationships and exhibit suboptimal performance in large-scale data scenarios.

%In the realm of time-series forecasting, statistical methodologies~\cite{kolkova2021demand} such as ARIMA~\cite{das2020learning} and Prophet~\cite{taylor2018forecasting}  have been widely utilized across various contexts. \textbf{ARIMA}, a classical model, employs differencing and moving averages within autoregression to adeptly address forecasting tasks, showcasing its effectiveness in predicting occupational task demands. Prophet decomposes time series into trend, seasonality, and holiday components, enabling it to handle both linear and nonlinear trends with changepoints. However, constrained by their linear or simple nonlinear structures, these models struggle to capture complex nonlinear relationships and exhibit suboptimal performance under large-scale data scenarios.

\textbf{RNN-based Model.}
RNN-based methods are effective in capturing temporal state transitions through their recurrent structures, making them widely used in various time series forecasting tasks~\cite{zhu2016recruitment,zhang2019aftershock,zhang2021talent,zhang2021exploiting,shen2019machine}. Notably, \textit{LSTM} have demonstrated their effectiveness in predicting changes in skill shares over time~\cite{garcia2022practical}. However, conventional RNNs often encounter performance degradation when handling excessively long look-back windows and forecast horizons. To address this challenge, \textit{SegRNN}~\cite{lin2023segrnn} introduces segment-wise iterations, which reduce the recurrence count within RNNs, thereby significantly enhancing performance in time series forecasting tasks.
%RNN-based methods, employing a recurrent structure, effectively capture temporal state transitions and have gained prominence in various applications, including those where \textbf{LSTM} and \textbf{GRU}~\cite{garcia2022practical} are utilized to predict skill shares over time. Nonetheless, the conventional RNN often encounters performance degradation due to excessively long look-back windows and forecast horizons. In response to this challenge, \textbf{SegRNN}~\cite{lin2023segrnn} introduces segment-wise iterations, reducing the recurrence count in RNNs and thereby circumventing the issue of extended look-back and forecast horizons.

\textbf{Transformer-based Model.}
Recently, Transformer-based models~\cite{vaswani2017attention} have gained widespread recognition in long-term time series forecasting due to their global modeling capabilities. Leveraging the attention mechanism, \textit{Reformer}~\cite{kitaev2020reformer} introduces locally sensitive hashing to approximate attention by grouping similar queries. \textit{Informer}~\cite{zhou2021informer} incorporates low-rank matrices in self-attention mechanisms to accelerate computation. \textit{Autoformer}\cite{wu2021autoformer} employs block decomposition and autocorrelation mechanisms to more effectively capture the intrinsic features of time series data. \textit{FedFormer}~\cite{zhou2022fedformer} utilizes DFT-based frequency-enhanced attention, obtaining attentive weights through the spectrums of queries and keys and calculating the weighted sum in the frequency domain. To address the challenges of non-stationary time series, the \textit{Non-stationary Transformer (NStransformer)}~\cite{liu2022non} introduces a sequence stabilization module and proposes a de-stationary attention mechanism. Additionally, \textit{PatchTST}~\cite{nie2022time} is a channel-independent patch time series transformer model that features patching and channel-independence as its key design elements.

%Recently, benefiting from the global modeling capacity, Transformer-based~\cite{vaswani2017attention} models have been widely acknowledged in long-term series forecasting. Focusing on the attention mechanism, \textbf{Reformer}~\cite{kitaev2020reformer} introduces locally sensitive hashing to approximate attention by assigning similar queries. 
%\textbf{Informer}~\cite{zhou2021informer} introduce the idea of low-rank matrices in self-attention mechanisms to accelerate the model’s computation process.
%\textbf{Autoformer}~\cite{wu2021autoformer} adopts block decomposition and autocorrelation to better capture the features of time series.
%\textbf{FedFormer}~\cite{zhou2022fedformer} proposes a DFT-based frequency enhanced attention, which obtains the attentive weights by the spectrums of queries and keys, and calculates the weighted sum in the frequency domain. 
%To solve challenges of non-stationary time series, \textbf{Non-stationary Transformer}~\cite{liu2022non} (NStransformer) firstly introduced a sequence stabilization module and proposed destationary attention mechanism. Furthermore, \textbf{PatchTST}~\cite{nie2022time} is a channel-independence patch time series transformer model that contains patching and channel-independence two key designs.

\textbf{MLP-based Model.}
Multiple Layer Projection (MLP) has been introduced in time series forecasting, demonstrating superior performance compared to transformer-based models in both accuracy and efficiency~\cite{zeng2023transformers}. Specifically, \textit{DLinear}~\cite{zeng2023transformers} uses series decomposition as a pre-processing step before linear regression. \textit{FreTS}~\cite{yi2024frequency} explores a novel approach by applying MLPs in the frequency domain for time series forecasting. \textit{TSMixer}~\cite{wang2024timemixer} employs MLPMixer blocks, segments input time series into fixed windows, and applies gated MLP transformations and permutations to enhance accuracy.
%Additionally, Multiple Layer Projection (MLP) has been introduced in time series forecasting, exhibiting favorable performance compared to transformer-based models in both forecasting accuracy and efficiency~\cite{zeng2023transformers}. Specifically, \textbf{DLinear}~\cite{zeng2023transformers} utilizes the series decomposition as the pre-processing before linear regression. \textbf{FreTS.}~\cite{yi2024frequency} explore a novel direction of applying MLPs in the frequency domain for time series forecasting. \textbf{TSMixer}~\cite{wang2024timemixer} employs MLPMixer blocks and segments input time series into fixed windows, followed by gated MLP transformations and permutations for enhanced accuracy. 

\textbf{Graph-based Models.}
Graph Neural Networks (GNNs) can learn non-Euclidean relationships, making them effective for identifying associations in structured data and generating joint representations from different perspectives~\cite{chen2022multi,chen2023entity,shen2021regularizing,chen2021trend}. CHGH~\cite{dl::skill-demand-supply}  uses an adaptive graph enhanced by skill co-occurrence relationships to link skill supply and demand sequences. This fusion of representations across views improves the performance of joint skill supply and demand prediction tasks.
Pre-DyGAE~\cite{chen2024pre} targets skill demand prediction from an occupational perspective. It builds an occupation-skill bipartite graph based on the skill demands of occupations and captures the dynamic changes in these relationships. This method allows for predicting both potential occupational skills and skill demands, leveraging a dynamic graph perspective.

\textbf{Fourier-based Models.}
By utilizing Fourier projection, \textit{FiLM}~\cite{zhou2022film} not only captures long-term time dependencies but also effectively reduces noise in forecasting. To address the challenge of non-stationary time-series forecasting, \textit{Koopa}~\cite{liu2024koopa} disentangles time-variant and time-invariant components from complex non-stationary series using a Fourier Filter and designs the Koopman Predictor to forecast dynamics.

%Nevertheless, there remains significant potential for improvement in preserving historical information in neural networks while mitigating overfitting to noise present in the data. Utilizing Fourier projection to mitigate noise, \textbf{FiLM}~\cite{zhou2022film} not only capture the long-term time dependencies but also effectively reduce the noise in forecasting. 
%Addressing the challenge of non-stationary time-series forecasting, \textbf{Koopa}~\cite{liu2024koopa}, disentangles time-variant and time-invariant components from intricate non-stationary series using Fourier Filter and designs the Koopman Predictor to predict dynamics.

% \subsection{Dataset Split}
% For the dataset split, we follow a standard protocol~\cite{MTS::wang2022learning} that categorizes all datasets ranging from January 2021 to December 2023 into training, validation, and test set in chronological order by the ratio of 9:1:2. Furthermore, the lookback window size was set to 6 months, indicating that the model considers the previous 6 months of data to make predictions. Furthermore, the prediction length was set to 3 months, signifying that the model forecasts skill demand for the subsequent 3 months based on the input data.

\subsection{Evaluation Metrics}
To evaluate the performance of various benchmark models in job skill demand forecasting tasks, we selected two commonly used regression metrics: MAE and RMSE. MAE is calculated over $H$ observations using the formula: $\frac{1}{H} \sum_{i=1}^{H}|y_i-\hat{y}_i|$, where $y_i$ represents the ground truth value and $\hat{y}_i$ is the predicted value. RMSE is calculated as: $\sqrt{\frac{1}{H} \sum_{i=1}^H\left(y_i-\hat{y}_i\right)^2}$. Both MAE and RMSE are scale-dependent metrics, which makes them unsuitable for comparison across different granularities. Additionally, these metrics are less sensitive to prediction errors at lower skill demand values. Therefore, we additionally applied SMAPE~\cite{sareminia2023support} and RRMSE~\cite{rawal2024mining} to assess the performance of various forecasting models. SMAPE considers both the magnitude and direction of errors, making it suitable for comparing forecasts across different scales. RRMSE measures the square root of the average of the squared percentage errors.
\begin{equation}
SMAPE = \frac{2}{H} * \sum_{i=1}^H \frac{\left|y_i-\hat{y}_i\right|}{\left|y_i\right|+\left|\hat{y}_i\right|}, \ \ \ RRMSE = \sqrt{\frac{\frac{1}{H} \sum_{i=1}^H\left(y_i-\hat{y}_i\right)^2}{\sum_{i=1}^H\left(\hat{y}_i\right)^2}}.
\end{equation}

% It is defined as: 

%RMSE is calculated over $H$ observations using the formula: $\sqrt{\frac{1}{H} \sum_{i=1}^H\left(y_i-\hat{y}_i\right)^2}$, where $y_i$ represents the ground truth value and $\hat{y}_i$ is the predicted value. 

%SMAPE considers both the magnitude and direction of errors. It is defined as: $\frac{2}{H} * \sum_{i=1}^H \frac{\left|y_i-\hat{y}_i\right|}{\left|y_i\right|+\left|\hat{y}_i\right|}$.

\begin{table}[t]
    \centering
    \caption{Performance comparison on MAE and RMSE.}
\resizebox{\textwidth}{!}{
\begin{tabular}{l|p{1cm}p{1cm}|p{1cm}p{1cm}|p{1cm}p{1cm}|p{1cm}p{1cm}|p{1cm}p{1cm}}
\toprule
\multirow{2}{*}{Model}&\multicolumn{2}{c|}{L1-Occupation}&\multicolumn{2}{c|}{L2-Occupation}&\multicolumn{2}{c|}{Region\&L1-O}&\multicolumn{2}{c|}{Region\&L2-O}&\multicolumn{2}{c}{Company}\\
% \hline
&MAE&RMSE&MAE&RMSE&MAE&RMSE&MAE&RMSE&MAE&RMSE\\
\midrule
ARIMA&20.27&256.89&6.46&115.79&3.98&58.65&1.31&27.42&1.31&38.88\\
Prophet&29.15&356.67&8.95&161.01&5.08&72.21&1.62&33.02&1.55&41.19\\
\midrule
LSTM&19.05&194.67&7.09&116.36&3.92&51.59&1.29&23.31&1.35&26.47\\
% GRU&19.27&195.75&7.16&117.12&3.98&51.70&1.30&23.37&1.36&26.64\\
SegRNN&12.28&108.28&5.01&68.83&3.14&34.26&1.05&15.96&1.01&16.03\\
\midrule
CHGH&22.09&261.49&7.09&116.58&3.91&51.46&1.28&23.24&1.34&26.52\\
Pre-DyGAE&22.98&187.90&7.04&82.97&4.24&38.62&1.37&17.39&1.24&18.24\\
\midrule
Transformer&22.06&215.09&7.58&118.21&4.01&52.04&1.35&23.44&1.26&24.99\\
Autoformer&23.06&186.76&8.22&100.02&6.45&57.77&2.41&24.10&3.31&38.55\\
Informer&22.21&205.24&7.43&117.38&3.88&50.13&1.30&23.07&1.26&24.92\\
Reformer&22.11&204.35&7.46&116.60&3.91&50.95&1.25&22.81&1.54&27.37\\
FEDformer&22.87&181.93&7.46&88.97&4.63&43.21&1.98&21.73&2.43&26.92\\
NStransformer&17.36&149.46&5.75&86.24&3.45&37.09&1.15&17.45&2.13&34.83\\
PatchTST&14.91&141.06&5.15&78.86&3.10&35.38&1.04&16.57&1.01&19.09\\
\midrule
DLinear&16.61&154.88&5.44&81.61&3.24&36.67&1.07&16.79&1.05&18.85\\
TSMixer&21.34&192.85&8.14&106.65&5.81&62.14&5.95&68.26&13.96&144.96\\
FreTS&16.47&167.61&6.52&106.39&3.65&47.81&1.22&21.92&1.26&25.39\\
\midrule
FiLM&12.95&117.17&5.08&65.65&3.24&29.90&1.14&14.01&1.17&15.87\\
Koopa&19.91&179.30&6.05&91.87&3.53&40.73&1.15&18.71&1.08&20.18\\
\bottomrule
\end{tabular}
}
\label{tab:overall}
\vspace{-5mm}
\end{table}

\subsection{Benchmark Results}

\textbf{Overall Performance.}
In Table~\ref{tab:overall}, we present the performance of various models evaluated using two metrics: MAE and RMSE. The following conclusions can be drawn: (1)~The traditional statistical method, Prophet, demonstrates relatively poor predictive performance. This may be due to seasonal and holiday factors not being the primary influencers in skill demand prediction. (2)~Most Transformer-based models, including Transformer, Autoformer, Informer, and Reformer, exhibit subpar overall predictive performance. This is likely because these models are designed to address long-range temporal dependencies, which are not well-suited for the current shorter time series context. (3)~In contrast, PatchTST, unlike these Transformer-based models that perform point-wise modeling of time series, segments the time series into patches and inputs them into the Transformer. This allows the model to focus on more local information. A similar idea is also employed in the SegRNN. This strategy significantly enhances the performance of these models in predicting job skill demand. (4)~The performance of different linear models on our dataset varies significantly. For instance, DLinear outperforms most Transformer-based models, while TSMixer performs poorly. This discrepancy may be due to the tendency of more complex MLP-based models to overfit our dataset. (5) CHGH and Pre-DyGAE exhibit poor performance in the separate skill demand forecasting scenario, likely due to a mismatch between their model design and the context of our dataset. Specifically, CHGH relies on sequential data from the supply side of skills, which is lacking in our dataset. Conversely, Pre-DyGAE focuses more on predicting whether a skill will be required by an occupation in the future. (6) Finally, FiLM achieved the best performance in most cases, demonstrating the robustness of the denoising-based model.

\textbf{Low-Demand Skill Prediction Performance.}
Considering the varying nature of skill demand values, we further employed SMAPE and RRMSE metrics to focus on the predictive performance of different models for low-demand skills. As shown in Table~\ref{tab:my_label}, the experimental results indicate the following: (1)~PatchTST achieved the best SMAPE performance in most cases, validating its ability to more accurately predict the trends of low-demand skills. (2)~Based on scale-independent metrics, we can compare the performance of models at different granularities. It can be observed that RRMSE exhibits a significant trend of variation across different granularities; specifically, as the granularity becomes finer, the RRMSE performance deteriorates. This indicates that predicting skill demand at finer granularities is more challenging. Additionally, FiLM shows the least variation across multiple granularities, further validating its ability to provide stable and reliable predictions under varying granularities and demand value ranges. (3)~Although Koopa performs averagely on MAE and RMSE metrics, it excels in predicting low-demand skills, particularly in terms of SMAPE. Similarly, NStransformer also performs well in scenarios focusing on low-demand skill predictions. This success can be attributed to both methods being designed to handle non-stationary time series. They effectively filter noise from historical sequences and restore intrinsic non-stationary information into time-dependent relationships, making them more adept at handling the fluctuating nature of low-demand skill time series data.

%The metrics RMSE and MAE, as they compute the absolute gap between predicted and actual values, primarily emphasize the prediction of values with significant errors, failing to reflect the accuracy of low-demand skill predictions. To address this issue, we further measure models's effectiveness using SMAPE and RRMSE, yielding the following conclusions.
%Unlike other methods, the performance of RRMSE worsens as granularity increases. However, the variation in RRMSE across different granularities for FiLM is not significant, demonstrating its accuracy and robustness in predicting low-demand skill requirements. Additionally, Nonstationary Transformer performs excellently across all metrics, indicating its ability to effectively capture the fluctuations in low-demand skill distributions and achieve strong predictive performance.

\begin{table}[t]
    \centering
    \caption{Performance comparison on SMAPE and RRMSE.}
\resizebox{\textwidth}{!}{
\begin{tabular}{l|p{1cm}p{1cm}|p{1cm}p{1cm}|p{1cm}p{1cm}|p{1cm}p{1cm}|p{1cm}p{1cm}}
\toprule
\multirow{2}{*}{Model}&\multicolumn{2}{c|}{L1-Occupation~(\%)}&\multicolumn{2}{c|}{L2-Occupation~(\%)}&\multicolumn{2}{c|}{Region\&L1-O~(\%)}&\multicolumn{2}{c|}{Region\&L2-O~(\%)}&\multicolumn{2}{c}{Company~(\%)}\\
% \hline
&SMAPE&RRMSE&SMAPE&RRMSE&SMAPE&RRMSE&SMAPE&RRMSE&SMAPE&RRMSE\\
\midrule
ARIMA&35.72&47.89&25.00&58.87&23.86&58.07&13.58&73.57&20.17&147.94\\
Prophet&41.22&67.78&28.35&88.47&26.75&71.60&15.07&93.04&22.31&167.77\\
\midrule
LSTM&41.38&57.90&32.85&83.70&31.58&68.40&22.93&87.36&30.26&174.40\\
% GRU&44.88&59.29&39.84&83.03&38.95&68.25&32.67&87.44&38.26&174.82\\
SegRNN&39.81&37.58&33.35&50.53&35.30&48.53&23.84&61.90&33.07&86.27\\
\midrule
CHGH&40.27&66.05&29.60&84.10&28.11&68.42&17.42&87.45&26.72&176.70\\
PreDyGAE&49.87&83.67&60.54&83.60&59.32&66.56&72.67&98.09&26.21&145.73\\
\midrule
Transformer&55.59&64.25&44.23&84.27&31.15&76.16&33.04&86.87&27.61&164.36\\
Autoformer&70.28&53.75&74.37&63.40&90.14&65.57&91.51&74.46&107.05&99.60\\
Informer&56.85&58.18&44.04&88.72&34.75&69.59&29.29&90.15&32.41&164.37\\
Reformer&56.58&61.35&40.58&83.70&32.21&72.87&20.86&90.85&45.25&169.87\\
FEDformer&69.30&54.03&69.29&60.00&73.17&52.69&81.73&70.06&94.19&97.97\\
NStransformer&38.11&47.19&26.30&60.73&24.98&48.89&14.55&63.29&24.20&100.78\\
PatchTST&34.70&51.17&24.52&58.80&25.15&44.96&13.50&67.48&19.89&115.34\\
\midrule
DLinear&41.84&52.89&34.35&60.22&33.47&51.05&25.77&64.65&30.71&108.66\\
TSMixer&56.59&61.17&72.29&99.35&82.48&87.29&120.85&96.49&155.20&102.14\\
FreTS&39.76&54.42&30.18&80.44&28.58&66.11&17.62&85.04&27.24&174.56\\
\midrule
FiLM&39.51&37.55&29.65&43.86&28.79&37.66&17.24&47.75&25.72&76.92\\
Koopa&37.84&58.30&25.72&65.34&24.41&57.81&13.98&74.00&20.43&123.96\\
\bottomrule
\end{tabular}
}
\label{tab:my_label}
\end{table}

\begin{table}[t]
    \centering
    \caption{Performance comparison on data with structural breaks on MAE and RMSE.}
\resizebox{\textwidth}{!}{
\begin{tabular}{l|p{1cm}p{1cm}|p{1cm}p{1cm}|p{1cm}p{1cm}|p{1cm}p{1cm}|p{1cm}p{1cm}}
\toprule
\multirow{2}{*}{Model}&\multicolumn{2}{c|}{L1-Occupation}&\multicolumn{2}{c|}{L2-Occupation}&\multicolumn{2}{c|}{Region\&L1-O}&\multicolumn{2}{c|}{Region\&L2-O}&\multicolumn{2}{c}{Company}\\
% \hline
&MAE&RMSE&MAE&RMSE&MAE&RMSE&MAE&RMSE&MAE&RMSE\\
% \midrule
% ARIMA&79.26&537.52&4.99&37.01&6.99&121.64&1.34&18.40&1.55&37.72\\
% prophet&117.45&746.99&6.91&60.48&8.61&133.56&1.74&26.54&1.75&37.08\\
\midrule
LSTM&87.30&554.46&57.95&400.22&18.99&149.53&7.91&52.38&24.40&159.02\\
% GRU&88.17&560.44&58.76&401.33&19.19&149.69&8.06&54.44&24.57&159.99\\
SegRNN&61.92&390.54&43.97&276.57&15.85&114.04&6.56&37.84&17.98&112.13\\
\midrule
CHGH&94.30&629.32&58.06&401.45&19.00&149.75&7.90&52.50&24.37&159.44\\
PreGyGAE&78.35&493.83&48.69&336.15&17.49&136.66&7.31&38.88&19.76&164.43\\
\midrule
Transformer&98.66&580.58&61.73&404.17&19.37&151.12&8.45&55.46&22.41&152.27\\
Autoformer&107.22&533.06&67.66&350.97&26.84&156.50&12.19&63.04&44.10&208.96\\
Informer&98.89&570.35&59.95&402.75&19.03&146.91&7.72&49.15&22.37&151.87\\
Reformer&98.14&569.83&60.71&401.21&19.25&149.91&7.52&49.10&25.65&160.69\\
FEDformer&105.43&532.24&62.10&325.10&20.49&128.45&10.37&55.47&34.09&155.28\\
NStransformer&82.43&462.24&49.30&318.44&16.59&119.91&6.85&37.56&40.05&196.03\\
PatchTST&77.44&474.86&45.02&303.76&14.88&111.01&6.56&38.60&18.03&127.72\\
\midrule
DLinear&81.17&485.25&46.67&307.34&15.94&118.94&6.50&37.72&18.18&124.32\\
TSMixer&107.47&614.93&83.60&479.39&29.99&187.08&25.83&190.29&155.10&766.58\\
FreTS&82.45&537.12&56.54&393.38&18.55&148.33&7.88&52.87&24.21&160.01\\
\midrule
FiLM&62.86&404.82&42.63&260.99&14.31&101.23&6.37&32.28&18.78&110.65\\
Koopa&91.26&516.75&50.44&324.15&17.43&128.39&7.07&41.29&19.04&133.26\\
\bottomrule
\end{tabular}
}
\label{tab:strcu1}
\vspace{-5mm}
\end{table}

\begin{table}[t]
    \centering
    \caption{Performance comparison on data with structural breaks on RRMSE and SMAPE.}
\resizebox{\textwidth}{!}{
\begin{tabular}{l|p{1cm}p{1cm}|p{1cm}p{1cm}|p{1cm}p{1cm}|p{1cm}p{1cm}|p{1cm}p{1cm}}
\toprule
\multirow{2}{*}{Model}&\multicolumn{2}{c|}{L1-Occupation~(\%)}&\multicolumn{2}{c|}{L2-Occupation~(\%)}&\multicolumn{2}{c|}{Region\&L1-O~(\%)}&\multicolumn{2}{c|}{Region\&L2-O~(\%)}&\multicolumn{2}{c}{Company~(\%)}\\
% \hline
&SMAPE&RRMSE&SMAPE&RRMSE&SMAPE&RRMSE&SMAPE&RRMSE&SMAPE&RRMSE\\
% \midrule
% ARIMA&42.43&48.19&24.26&23.47&24.93&84.82&11.96&29.06&19.44&100.41\\
% prophet&57.04&68.33&27.23&41.35&27.75&99.48&13.19&46.92&21.32&102.51\\
\midrule
LSTM&43.78&58.05&48.93&84.46&46.64&78.31&42.03&58.48&68.38&187.30\\
% GRU&44.33&59.48&49.54&83.83&49.39&77.53&46.08&60.60&68.72&188.06\\
SegRNN&39.22&37.80&43.09&51.14&45.17&54.31&39.41&41.40&57.45&89.65\\
\midrule
CHGH&44.91&66.31&48.90&84.87&45.43&78.32&39.79&58.89&68.36&189.91\\
PreDyGAE&52.35&47.15&56.56&59.31&52.06&61.22&44.13&42.31&70.26&106.88\\
\midrule
Transformer&50.01&64.47&53.10&84.95&46.50&86.56&47.67&61.23&64.92&177.43\\
Autoformer&63.46&54.08&68.62&64.14&87.93&68.97&88.95&63.85&115.00&100.60\\
Informer&51.11&58.40&51.89&89.70&47.81&80.86&44.90&57.55&65.11&177.16\\
Reformer&50.79&61.59&51.51&84.53&46.86&84.15&40.81&58.59&72.36&181.36\\
FEDformer&62.83&54.37&64.37&60.84&72.24&58.55&80.03&54.29&103.27&100.65\\
NStransformer&45.36&47.46&47.63&61.85&43.04&57.60&36.72&39.72&170.57&113.87\\
PatchTST&40.89&51.48&43.26&59.69&41.51&51.85&34.74&43.12&55.26&122.56\\
\midrule
DLinear&43.14&53.20&45.25&61.13&45.26&58.80&41.15&41.71&57.65&115.24\\
TSMixer&54.31&61.31&76.08&99.84&85.12&95.81&117.39&93.66&160.55&102.23\\
FreTS&42.44&54.59&48.24&81.17&45.39&75.43&39.85&57.83&68.39&187.94\\
\midrule
FiLM&38.96&37.82&44.23&44.52&44.95&43.06&40.05&30.80&56.37&80.77\\
Koopa&46.45&58.59&47.13&66.28&42.60&66.20&36.24&47.48&58.98&131.77\\
\bottomrule
\end{tabular}
}
\label{tab:strcu2}
\vspace{-5mm}
\end{table}

\textbf{Performance on Skill Demand Series with Structural Breaks.}
As described in Section 3.3, in the dynamically changing job market, skill demand time series data exhibit structural breaks. To assess the impact of this phenomenon on different models in the skill demand forecasting task, we used the Chow test to detect structural breaks in the skill demand time series. The corresponding predictive performance of different models is presented in Tables~\ref{tab:strcu1} and \ref{tab:strcu2}. We observe the following phenomena: (1)~Compared to the predictive performance on the full dataset, the performance on time series data with structural breaks is significantly worse. This finding underscores the complexity and unpredictability of skill trends that experience structural breaks. (2)~FiLM has achieved results close to the overall skill demand prediction in terms of SMAPE and RRMSE metrics. This validates that FiLM can effectively mitigate the disruptive impact of structural breaks on skill demand forecasting. (3)~Furthermore, while the overall predictive performance of skill demand forecasting at both the Region\&L2-O and Company granularity levels is similar, significant differences emerge when forecasting skills experiencing structural breaks. This suggests that skills undergoing structural breaks display more predictable patterns at the Region\&L2-O granularity level compared to the Company level, making them relatively easier to forecast.

%We augment our experiments to focus on the evaluation of time series related to structural breaks. Given the rapid evolution of the labor market, leading to swift changes in skill demands, we aim to scrutinize the predictive outcomes concerning skills that have undergone structural breaks. Thus, we employ the Chow test~\cite{chow1960tests} to filter skill demand series for the test set, subsequently observing the effects of existing models on skill prediction under structural breaks.
%As shown in Table \ref{tab:strcu2} owing to structural breaks, there is a significant decrease in overall metrics such as RMSE, indicating the challenge in forecasting these skill demands. Subsequently, in the prediction of such skill demands, Nonstationary Transformer and SegRNN exhibit the most favorable overall performance, showcasing an ability to discern the occurrence of structural mutations effectively.

%Our experimental analysis provides insights into the performance of various models in predicting skill demand. These findings offer valuable insights for selecting suitable models tailored to specific task requirements and understanding the strengths and limitations of different approaches in skill demand forecasting.

\section{Conclusion}
In this work, we introduced Job-SDF, a dataset designed for training and benchmarking job-skill demand forecasting models. Compiled from millions of public job advertisements collected from online recruitment platforms, this dataset includes monthly recruitment demand for 2,324 types of skills across 52 occupations, 521 companies, and 7 regions. Using this dataset, we validated a wide range of time-series forecasting approaches, including statistical models, RNN-based models, Transformer-based models, MLP-based models, Graph-based models, and Fourier-based models. Furthermore, we conducted extensive experiments to compare the performance of various methods in predicting skill demand at different granularities. We hope that Job-SDF will facilitate further research in this field.

%In this work, we have developed the Job-SDF dataset along with corresponding benchmark methods and frameworks. The dataset is continuously updated and accessible online, encompassing skill demands at various granularities from January 2021 to December 2023, spanning companies, occupations, and regions. Leveraging this dataset, we have conducted validation across a plethora of time-series forecasting models, including statistical methods, RNN-based, Transformer-based, Graph-based, and Fourier-based. Our experimental findings illuminate the unique characteristics of our dataset: intricate skill interdependencies, a long-tailed distribution of skill demands, substantial variations in demand across different granularities, among others, underscoring the necessity of our proposed dataset and benchmarking methodologies. Ultimately, we introduce the inaugural publicly available dataset for skill demand prediction, promising substantial advancements in research within this domain.

\section*{Acknowledgements}
This work was supported in part by the National Key R\&D Program of China (Grant No.2023YFF0725001), in part by the National Natural Science Foundation of China (Grant No.92370204), in part by the guangdong Basic and Applied Basic Research Foundation (Grant No.2023B1515120057), in part by Guangzhou-HKUST (GZ) Joint Funding Program (Grant No.2023A03J0008), Education Bureau of Guangzhou Municipality, in part by Nansha Postdoctoral Research Project, and in part by the National Natural Science Foundation of China (Grant No.62176014), the Fundamental Research Funds for the Central Universities.

\bibliographystyle{unsrt}
\bibliography{reference}

%%%%%%%%%%%%%%%%%%%%%%%%%%%%%%%%%%%%%%%%%%%%%%%%%%%%%%%%%%%%
%%%%%%%%%%%%%%%%%%%%%%%%%%%%%%%%%%%%%%%%%%%%%%%%%%%%%%%%%%%%
\newpage
% \includeonly{Computation Resource} 
% \renewcommand{\contentsname}{Appendix}
% \tableofcontents
\newpage
\appendix

\section{Computational Resource}
Due to inherent design and size constraints of the models combined with varying data sizes at different granularities, the deployment environments for each model are distinct. The CHGH model, which requires over 80GB of memory, is exclusively deployed on CPU platforms to accommodate its substantial resource demands. In contrast, the PreDyGAE model operates solely on GPU infrastructure, leveraging the computational efficiencies of the NVIDIA A800 GPUs.
For other models, deployment strategies are tailored according to the granularity of the data. Experiments at the labor market, regions, L1 occupations, L2 occupations, and Region \& L1 occupations granularities are conducted on GPUs, capitalizing on the enhanced processing capabilities of these units for handling moderate data volumes. However, at the granularities of Region \& L2 and company, where data volumes are significantly larger, deployment shifts to CPUs.
Overall, the training time of different models are shown in Table \ref{tab:time}.

\begin{table}[ht!]
    \caption{Training time (minute) of different models for job skill demand forecsting.}
    \centering
\resizebox{\textwidth}{!}{
\begin{tabular}{l|p{1.5cm}p{1.5cm}p{1.5cm}p{1.5cm}p{1.5cm}p{1.5cm}p{1.5cm}}
\toprule
Model&Market&Region&L1-O&L2-O&R\&L1-O&R\&L2-O&Company\\
\midrule
LSTM&0-0.5&0-0.5&0-0.5&0-0.5&0-0.5&37.7&39.0\\
SegRNN&0-0.5&0-0.5&0-0.5&0-0.5&0-0.5&342.8&458.2\\
CHGH&17.7&132.8&170.2&258.3&1300.3&490.6&6604.2\\
PreDyGAE&1-10&16.5&30.0&48.1&48.1&88.2&126.2\\
Transformer&0-0.5&0-0.5&0-0.5&0.5-1&0.5-1&128.2&166.5\\
Autoformer&0-0.5&0-0.5&0-0.5&0-0.5&0.5-1&304.3&325.0\\
Informer&0-0.5&0-0.5&0-0.5&0-0.5&0-0.5&133.8&171.7\\
Reformer&0-0.5&0-0.5&0-0.5&0-0.5&0-0.5&36.0&52.5\\
FEDformer&0-0.5&0-0.5&0-0.5&0-0.5&0-0.5&193.8&198.5\\
NStransformer&0-0.5&0.5-1&0.5-1&0-0.5&0.5-1&128.5&195.7\\
PatchTST&0-0.5&0-0.5&0-0.5&0-0.5&0.5-1&1202.8&2558.0\\
DLinear&0-0.5&0-0.5&0.5-1&0-0.5&0-0.5&20.0&39.1\\
TSMixer&0-0.5&0-0.5&0-0.5&0-0.5&0-0.5&24.0&97.0\\
FreTS&0-0.5&0-0.5&0-0.5&0-0.5&0-0.5&85.0&200.0\\
FiLM&0-0.5&0.5-1&0.5-1&1-10&1-10&598.0&1464.7\\
Koopa&0-0.5&0-0.5&0-0.5&0-0.5&0-0.5&38.3&68.5\\
\bottomrule
\end{tabular}
}
    \label{tab:time}
\end{table}

\section{Additional Experimental Results}
\label{app::add_exp}
Due to page limitations in the main text, we have included additional experimental content in the appendix. First, we present the results of repeated trials of the benchmark models discussed in the main text in the first subsection. Subsequently, we focus on the performance of existing benchmark models in predicting demand for low-frequency skills. Further, we have constructed a co-occurrence relationship between skills as prior knowledge based on the training set and employed various Graph Neural Network (GNN)-based multivariate time series forecasting methods in the task of job skill demand forecasting, demonstrating promising results. Finally, considering that the skill demand proportion may be more meaningful than the skill demand volume in certain contexts, we have constructed a dataset for skill demand proportion and showcased the performance of benchmark models on this task.

\subsection{Repeated Experiments on Job Skill Demand Forecasting}
To demonstrate the robustness and reliability of our experimental results, we first repeated the experiments multiple times as described in the main text. Additionally, we extended our analysis to include experiments across the entire labor market and at various regional granularities.

\paragraph{Implementment Details.}
% (How to choose the hyperparameters)
We utilized the Time-series-Library~\footnote{https://github.com/thuml/Time-Series-Library} to implement some of the models. The hyperparameters were uniformly set as follows: a learning rate of 0.0001, 20 epochs of training, a hidden layer dimension of 2048, the GELU activation function, and MSE loss as the loss function. Early stopping was employed to prevent overfitting by terminating training early when necessary.
Data from the first 24 months were used for pre-training, and the model was fine-tuned using the next 6 months to capture trend changes. Finally, the model was used to infer skill demands for the last 6 months. All other hyperparameters were kept consistent with those in the original paper.
To ensure the reliability of our findings, we repeated these experiments four times, using random seeds set to 0, 1, 2, and 3, respectively.

\paragraph{Overall Performance}
Table \ref{tab:overall_app} displays the mean and standard deviation results of repeated experiments on the benchmark models for the job skill demand forecasting task as presented in the main text. Initially, we supplement the experimental results at the overall labor market level and regional granularity, where the RMSE averages over 1000. In cases of coarser granularity, due to the larger base of demand values, the prediction deviations are significant. 
% Furthermore, the results of the repeated experiments are consistent with those of the main text experiments.

\begin{table}[ht!]
    \centering
    \caption{Overall performance comparisons on repeated experiments.}
    \label{tab:overall_app}
\resizebox{\textwidth}{!}{
\begin{tabular}{l|llllllll}
\toprule
&Model&Market&Region&L1-O&L2-O&R\&L1-O&R\&L2-O&Company\\
\midrule
\multirow{16}{*}{\rotatebox{90}{MAE}}&LSTM&$314.54\pm_{0.57}$&$49.92\pm_{0.0}$&$24.43\pm_{4.91}$&$8.04\pm_{0.87}$&$4.44\pm_{0.47}$&$1.45\pm_{0.15}$&$1.49\pm_{0.13}$\\
&SegRNN&$190.05\pm_{0.37}$&$35.92\pm_{0.47}$&$16.37\pm_{3.73}$&$5.81\pm_{0.73}$&$3.68\pm_{0.5}$&$1.32\pm_{0.25}$&$1.23\pm_{0.2}$\\
&CHGH&$315.47\pm_{0.04}$&$50.03\pm_{0.02}$&$25.62\pm_{3.23}$&$8.04\pm_{0.87}$&$4.43\pm_{0.48}$&$1.51\pm_{0.7}$&$1.54\pm_{0.73}$\\
&PreDyGAE&$189.95\pm_{0.01}$&$35.84\pm_{0.09}$&$21.72\pm_{1.15}$&$6.81\pm_{0.21}$&$4.08\pm_{0.15}$&$1.85\pm_{0.43}$&$1.85\pm_{0.56}$\\
&Transformer&$340.72\pm_{0.79}$&$54.98\pm_{0.13}$&$27.43\pm_{4.9}$&$8.93\pm_{1.24}$&$4.92\pm_{0.83}$&$1.75\pm_{0.36}$&$1.69\pm_{0.39}$\\
&Autoformer&$465.86\pm_{2.78}$&$60.9\pm_{0.64}$&$31.97\pm_{8.13}$&$9.97\pm_{1.6}$&$6.68\pm_{0.21}$&$2.6\pm_{0.17}$&$2.85\pm_{0.42}$\\
&Informer&$340.76\pm_{1.6}$&$55.08\pm_{0.35}$&$27.54\pm_{4.87}$&$8.81\pm_{1.26}$&$4.87\pm_{0.91}$&$1.73\pm_{0.39}$&$1.69\pm_{0.39}$\\
&Reformer&$344.9\pm_{2.62}$&$54.83\pm_{0.2}$&$27.35\pm_{4.79}$&$8.88\pm_{1.3}$&$4.9\pm_{0.9}$&$1.71\pm_{0.42}$&$1.8\pm_{0.24}$\\
&FEDformer&$469.46\pm_{4.52}$&$60.37\pm_{0.19}$&$31.87\pm_{8.22}$&$9.54\pm_{1.9}$&$5.83\pm_{1.1}$&$2.41\pm_{0.39}$&$2.52\pm_{0.08}$\\
&NStransformer&$208.23\pm_{3.43}$&$37.39\pm_{0.46}$&$19.13\pm_{1.61}$&$6.29\pm_{0.5}$&$3.79\pm_{0.31}$&$1.26\pm_{0.1}$&$1.64\pm_{0.44}$\\
&PatchTST&$204.94\pm_{12.41}$&$36.33\pm_{2.23}$&$18.69\pm_{3.45}$&$6.12\pm_{0.89}$&$3.69\pm_{0.54}$&$1.23\pm_{0.18}$&$1.21\pm_{0.18}$\\
&DLinear&$201.05\pm_{1.95}$&$35.45\pm_{0.56}$&$18.23\pm_{1.48}$&$5.89\pm_{0.41}$&$3.55\pm_{0.28}$&$1.21\pm_{0.12}$&$1.18\pm_{0.11}$\\
&TSMixer&$517.95\pm_{1.49}$&$67.56\pm_{11.06}$&$30.83\pm_{8.66}$&$10.54\pm_{2.19}$&$5.95\pm_{0.12}$&$3.58\pm_{2.16}$&$6.75\pm_{6.58}$\\
&FreTS&$310.62\pm_{4.57}$&$49.52\pm_{0.73}$&$23.1\pm_{6.05}$&$7.74\pm_{1.11}$&$4.3\pm_{0.59}$&$1.41\pm_{0.17}$&$1.45\pm_{0.17}$\\
&FiLM&$201.7\pm_{24.86}$&$36.18\pm_{2.29}$&$16.79\pm_{3.51}$&$5.9\pm_{0.75}$&$3.7\pm_{0.42}$&$1.28\pm_{0.13}$&$1.31\pm_{0.12}$\\
&Koopa&$205.12\pm_{4.95}$&$35.95\pm_{0.5}$&$19.9\pm_{0.01}$&$6.19\pm_{0.13}$&$3.68\pm_{0.13}$&$1.2\pm_{0.05}$&$1.15\pm_{0.06}$\\
\midrule
\multirow{16}{*}{\rotatebox{90}{RMSE}}&LSTM&$1799.7\pm_{0.49}$&$341.97\pm_{0.07}$&$308.58\pm_{103.99}$&$148.32\pm_{29.18}$&$66.87\pm_{13.95}$&$30.01\pm_{6.11}$&$35.35\pm_{8.1}$\\
&SegRNN&$941.59\pm_{1.08}$&$194.73\pm_{1.59}$&$154.02\pm_{41.75}$&$78.84\pm_{9.14}$&$38.6\pm_{3.97}$&$18.21\pm_{2.05}$&$21.05\pm_{4.58}$\\
&CHGH&$1803.56\pm_{0.73}$&$344.15\pm_{0.11}$&$335.97\pm_{67.99}$&$148.94\pm_{29.54}$&$67.12\pm_{14.29}$&$19.3\pm_{12.73}$&$20.61\pm_{14.53}$\\
&PreDyGAE&$925.27\pm_{0.01}$&$193.94\pm_{0.04}$&$186.36\pm_{1.4}$&$84.55\pm_{1.44}$&$40.37\pm_{1.6}$&$18.79\pm_{1.27}$&$22.05\pm_{3.48}$\\
&Transformer&$1829.87\pm_{2.96}$&$360.87\pm_{6.13}$&$318.17\pm_{94.1}$&$151.16\pm_{30.08}$&$67.66\pm_{14.26}$&$30.42\pm_{6.37}$&$35.12\pm_{9.25}$\\
&Autoformer&$2177.05\pm_{87.58}$&$313.84\pm_{8.55}$&$270.26\pm_{76.23}$&$122.29\pm_{20.33}$&$62.56\pm_{4.38}$&$26.57\pm_{2.26}$&$35.47\pm_{2.81}$\\
&Informer&$1854.81\pm_{3.0}$&$359.97\pm_{7.72}$&$314.97\pm_{100.17}$&$148.73\pm_{28.62}$&$67.01\pm_{15.41}$&$30.14\pm_{6.45}$&$35.1\pm_{9.3}$\\
&Reformer&$1890.32\pm_{13.72}$&$357.61\pm_{1.23}$&$314.03\pm_{100.12}$&$152.43\pm_{32.71}$&$67.58\pm_{15.18}$&$30.21\pm_{6.76}$&$36.14\pm_{8.01}$\\
&FEDformer&$2137.44\pm_{58.69}$&$321.53\pm_{2.51}$&$268.73\pm_{79.24}$&$113.47\pm_{22.37}$&$56.36\pm_{12.0}$&$25.64\pm_{3.57}$&$30.9\pm_{3.64}$\\
&NStransformer&$1121.07\pm_{25.29}$&$236.56\pm_{12.38}$&$198.03\pm_{44.33}$&$103.2\pm_{15.48}$&$45.99\pm_{8.13}$&$21.42\pm_{3.62}$&$32.8\pm_{1.86}$\\
&PatchTST&$1098.82\pm_{43.54}$&$220.21\pm_{7.5}$&$204.56\pm_{57.97}$&$97.9\pm_{17.38}$&$45.32\pm_{9.07}$&$20.88\pm_{3.94}$&$25.48\pm_{5.83}$\\
&DLinear&$1107.0\pm_{28.41}$&$222.14\pm_{4.31}$&$211.02\pm_{51.25}$&$99.81\pm_{16.61}$&$46.04\pm_{8.55}$&$21.12\pm_{3.95}$&$25.7\pm_{6.25}$\\
&TSMixer&$2578.82\pm_{27.27}$&$438.06\pm_{67.68}$&$350.22\pm_{143.66}$&$172.63\pm_{60.23}$&$78.88\pm_{15.28}$&$48.76\pm_{17.8}$&$82.91\pm_{56.64}$\\
&FreTS&$1763.69\pm_{25.03}$&$336.65\pm_{4.04}$&$289.29\pm_{111.07}$&$140.54\pm_{31.17}$&$63.72\pm_{14.52}$&$28.68\pm_{6.17}$&$34.0\pm_{7.86}$\\
&FiLM&$1071.14\pm_{223.85}$&$216.77\pm_{38.28}$&$173.2\pm_{51.15}$&$83.31\pm_{16.12}$&$39.26\pm_{8.54}$&$18.23\pm_{3.85}$&$22.53\pm_{6.08}$\\
&Koopa&$1147.88\pm_{31.89}$&$228.36\pm_{7.6}$&$227.46\pm_{43.96}$&$106.03\pm_{12.93}$&$49.37\pm_{7.89}$&$22.53\pm_{3.49}$&$27.06\pm_{6.28}$\\
\midrule
\multirow{16}{*}{\rotatebox{90}{SMAPE(\%)}}&LSTM&$42.33\pm_{0.04}$&$49.78\pm_{0.03}$&$42.31\pm_{0.85}$&$33.14\pm_{0.27}$&$31.79\pm_{0.19}$&$22.67\pm_{0.24}$&$30.42\pm_{0.14}$\\
&SegRNN&$39.67\pm_{0.26}$&$51.14\pm_{0.53}$&$42.16\pm_{2.14}$&$34.62\pm_{1.16}$&$36.02\pm_{0.65}$&$30.62\pm_{6.19}$&$32.77\pm_{0.27}$\\
&CHGH&$42.21\pm_{0.01}$&$49.03\pm_{0.03}$&$40.83\pm_{0.51}$&$29.98\pm_{0.35}$&$28.42\pm_{0.28}$&$46.97\pm_{9.54}$&$30.69\pm_{14.64}$\\
&PreDyGAE&$42.1\pm_{0.14}$&$64.37\pm_{1.24}$&$75.3\pm_{23.21}$&$59.69\pm_{0.77}$&$46.51\pm_{11.69}$&$76.6\pm_{3.58}$&$60.36\pm_{31.18}$\\
&Transformer&$49.86\pm_{0.46}$&$60.59\pm_{0.02}$&$58.06\pm_{2.26}$&$52.7\pm_{7.73}$&$47.37\pm_{14.81}$&$47.03\pm_{12.77}$&$45.69\pm_{16.5}$\\
&Autoformer&$73.47\pm_{0.36}$&$79.08\pm_{0.19}$&$79.44\pm_{8.36}$&$79.16\pm_{4.38}$&$86.16\pm_{3.64}$&$86.32\pm_{4.73}$&$91.12\pm_{14.54}$\\
&Informer&$49.2\pm_{0.2}$&$60.45\pm_{0.03}$&$58.59\pm_{1.59}$&$52.61\pm_{7.82}$&$48.84\pm_{12.86}$&$45.52\pm_{14.82}$&$47.61\pm_{13.88}$\\
&Reformer&$48.69\pm_{0.17}$&$60.48\pm_{0.17}$&$58.49\pm_{1.74}$&$51.26\pm_{9.75}$&$47.8\pm_{14.24}$&$42.15\pm_{19.43}$&$52.76\pm_{6.85}$\\
&FEDformer&$73.57\pm_{0.53}$&$78.11\pm_{0.55}$&$79.26\pm_{9.09}$&$76.98\pm_{7.02}$&$78.54\pm_{4.9}$&$82.05\pm_{0.3}$&$86.29\pm_{7.21}$\\
&NStransformer&$38.88\pm_{0.09}$&$47.74\pm_{0.14}$&$38.94\pm_{0.76}$&$27.49\pm_{1.08}$&$26.43\pm_{1.33}$&$15.47\pm_{0.84}$&$23.75\pm_{0.41}$\\
&PatchTST&$37.56\pm_{2.15}$&$46.52\pm_{2.49}$&$37.82\pm_{2.85}$&$26.98\pm_{2.25}$&$26.7\pm_{1.42}$&$15.2\pm_{1.55}$&$22.29\pm_{2.19}$\\
&DLinear&$37.7\pm_{1.82}$&$49.27\pm_{3.85}$&$41.79\pm_{0.05}$&$34.65\pm_{0.28}$&$34.07\pm_{0.54}$&$26.6\pm_{0.75}$&$31.42\pm_{0.65}$\\
&TSMixer&$73.63\pm_{1.83}$&$72.27\pm_{9.41}$&$61.01\pm_{4.04}$&$68.37\pm_{3.58}$&$65.19\pm_{15.78}$&$79.28\pm_{37.95}$&$95.62\pm_{54.39}$\\
&FreTS&$42.81\pm_{0.56}$&$50.46\pm_{0.62}$&$41.53\pm_{1.62}$&$31.26\pm_{0.98}$&$29.76\pm_{1.08}$&$18.89\pm_{1.16}$&$28.37\pm_{1.03}$\\
&FiLM&$37.79\pm_{1.75}$&$46.85\pm_{4.01}$&$41.03\pm_{1.38}$&$30.26\pm_{0.56}$&$29.49\pm_{0.64}$&$17.65\pm_{0.37}$&$26.19\pm_{0.42}$\\
&Koopa&$36.26\pm_{0.39}$&$44.7\pm_{0.95}$&$36.69\pm_{1.05}$&$25.59\pm_{0.12}$&$24.51\pm_{0.09}$&$14.13\pm_{0.14}$&$20.62\pm_{0.18}$\\
\midrule
\multirow{16}{*}{\rotatebox{90}{RRMSE(\%)}}&LSTM&$52.7\pm_{0.09}$&$52.87\pm_{0.03}$&$73.45\pm_{14.2}$&$97.42\pm_{12.52}$&$80.24\pm_{10.81}$&$102.19\pm_{13.54}$&$210.9\pm_{33.32}$\\
&SegRNN&$20.55\pm_{0.05}$&$22.49\pm_{0.39}$&$32.2\pm_{4.91}$&$41.44\pm_{8.3}$&$39.72\pm_{8.04}$&$49.27\pm_{11.53}$&$71.42\pm_{13.56}$\\
&CHGH&$52.92\pm_{0.03}$&$53.46\pm_{0.02}$&$77.26\pm_{10.23}$&$98.13\pm_{12.8}$&$80.86\pm_{11.35}$&$54.98\pm_{47.9}$&$100.68\pm_{96.78}$\\
&PreDyGAE&$20.07\pm_{0.0}$&$22.65\pm_{0.01}$&$51.16\pm_{29.68}$&$54.87\pm_{26.22}$&$47.52\pm_{17.38}$&$64.98\pm_{30.23}$&$96.4\pm_{45.03}$\\
&Transformer&$52.49\pm_{0.18}$&$55.57\pm_{1.14}$&$75.98\pm_{10.71}$&$98.09\pm_{12.61}$&$82.63\pm_{5.9}$&$101.34\pm_{13.21}$&$200.07\pm_{32.6}$\\
&Autoformer&$49.81\pm_{1.73}$&$38.47\pm_{1.72}$&$54.53\pm_{0.71}$&$64.59\pm_{1.09}$&$61.08\pm_{4.1}$&$71.01\pm_{3.15}$&$103.89\pm_{3.92}$\\
&Informer&$54.11\pm_{0.2}$&$55.59\pm_{2.17}$&$71.42\pm_{12.08}$&$97.76\pm_{8.25}$&$81.04\pm_{10.45}$&$102.15\pm_{10.95}$&$198.51\pm_{31.17}$\\
&Reformer&$55.75\pm_{0.54}$&$54.83\pm_{0.62}$&$76.12\pm_{13.48}$&$98.47\pm_{13.48}$&$81.65\pm_{8.01}$&$103.29\pm_{11.35}$&$206.04\pm_{33.02}$\\
&FEDformer&$48.25\pm_{2.89}$&$39.48\pm_{0.85}$&$56.52\pm_{2.27}$&$60.81\pm_{0.74}$&$55.93\pm_{2.95}$&$69.07\pm_{0.9}$&$102.11\pm_{3.78}$\\
&NStransformer&$26.75\pm_{0.76}$&$30.42\pm_{1.94}$&$43.64\pm_{3.24}$&$59.15\pm_{1.44}$&$49.08\pm_{0.17}$&$62.42\pm_{0.79}$&$107.45\pm_{6.09}$\\
&PatchTST&$26.38\pm_{1.12}$&$28.31\pm_{1.01}$&$47.51\pm_{3.34}$&$56.73\pm_{1.89}$&$47.96\pm_{2.74}$&$64.14\pm_{3.05}$&$108.92\pm_{5.86}$\\
&DLinear&$26.6\pm_{1.35}$&$28.58\pm_{1.32}$&$49.04\pm_{3.52}$&$58.69\pm_{1.4}$&$51.47\pm_{0.38}$&$64.72\pm_{0.06}$&$110.8\pm_{1.95}$\\
&TSMixer&$74.69\pm_{3.7}$&$69.2\pm_{15.69}$&$86.56\pm_{23.18}$&$118.56\pm_{17.54}$&$99.17\pm_{10.85}$&$100.88\pm_{4.01}$&$136.0\pm_{30.91}$\\
&FreTS&$51.01\pm_{0.94}$&$51.53\pm_{0.82}$&$69.56\pm_{13.82}$&$92.39\pm_{10.91}$&$76.55\pm_{9.53}$&$97.37\pm_{11.25}$&$200.81\pm_{23.97}$\\
&FiLM&$25.72\pm_{7.38}$&$27.83\pm_{7.19}$&$36.13\pm_{1.29}$&$43.32\pm_{0.5}$&$38.93\pm_{1.16}$&$48.51\pm_{0.7}$&$79.66\pm_{2.5}$\\
&Koopa&$27.81\pm_{1.3}$&$29.66\pm_{1.55}$&$52.74\pm_{5.07}$&$62.29\pm_{2.78}$&$56.41\pm_{1.28}$&$70.91\pm_{2.82}$&$122.3\pm_{1.52}$\\
\bottomrule
\end{tabular}

}
    
\end{table}

\paragraph{Performance on Skill Demand Series with Structural Breaks}

\begin{table}[ht!]
    \centering
    \caption{Performance comparisons on skill demand series with structural breaks.}
\resizebox{\textwidth}{!}{
\begin{tabular}{l|llllllll}
\toprule
&Model&Market&Region&L1-O&L2-O&R\&L1-O&R\&L2-O&Company\\
\midrule
\multirow{16}{*}{\rotatebox{90}{MAE}}&LSTM&$423.99\pm_{0.56}$&$109.63\pm_{0.04}$&$101.68\pm_{13.13}$&$63.39\pm_{4.97}$&$21.02\pm_{1.85}$&$8.55\pm_{0.59}$&$26.23\pm_{1.67}$\\
&SegRNN&$256.67\pm_{0.54}$&$76.09\pm_{1.01}$&$68.08\pm_{5.62}$&$44.93\pm_{0.88}$&$16.12\pm_{0.24}$&$7.07\pm_{0.47}$&$18.91\pm_{0.85}$\\
&CHGH&$425.17\pm_{0.04}$&$109.82\pm_{0.06}$&$104.41\pm_{9.23}$&$63.54\pm_{5.0}$&$18.11\pm_{1.93}$&$15.16\pm_{4.33}$&$19.75\pm_{13.35}$\\
&PreDyGAE&$296.27\pm_{0.01}$&$85.05\pm_{5.06}$&$84.11\pm_{8.87}$&$56.28\pm_{7.2}$&$16.57\pm_{0.84}$&$9.51\pm_{0.19}$&$22.9\pm_{0.13}$\\
&Transformer&$460.45\pm_{2.31}$&$120.62\pm_{0.4}$&$113.11\pm_{13.19}$&$69.45\pm_{7.05}$&$22.4\pm_{2.76}$&$9.42\pm_{0.88}$&$26.93\pm_{4.12}$\\
&Autoformer&$632.29\pm_{2.76}$&$131.92\pm_{1.52}$&$131.01\pm_{21.72}$&$75.4\pm_{7.07}$&$25.82\pm_{0.93}$&$11.76\pm_{0.39}$&$36.65\pm_{6.8}$\\
&Informer&$462.88\pm_{1.99}$&$120.89\pm_{0.68}$&$113.36\pm_{13.21}$&$68.13\pm_{7.46}$&$22.22\pm_{2.91}$&$9.1\pm_{1.26}$&$26.82\pm_{4.06}$\\
&Reformer&$465.72\pm_{3.72}$&$120.58\pm_{0.34}$&$112.51\pm_{13.12}$&$69.02\pm_{7.59}$&$22.4\pm_{2.88}$&$9.09\pm_{1.44}$&$28.16\pm_{2.29}$\\
&FEDformer&$645.02\pm_{6.45}$&$131.05\pm_{0.27}$&$130.18\pm_{22.59}$&$72.12\pm_{9.14}$&$23.06\pm_{2.34}$&$11.01\pm_{0.59}$&$32.89\pm_{1.1}$\\
&NStransformer&$285.18\pm_{6.1}$&$81.54\pm_{1.2}$&$80.2\pm_{2.03}$&$49.66\pm_{0.33}$&$16.88\pm_{0.26}$&$7.07\pm_{0.2}$&$28.34\pm_{10.69}$\\
&PatchTST&$282.34\pm_{17.19}$&$79.1\pm_{4.75}$&$80.46\pm_{2.76}$&$48.45\pm_{3.13}$&$16.29\pm_{1.29}$&$7.01\pm_{0.41}$&$19.65\pm_{1.48}$\\
&DLinear&$277.33\pm_{2.32}$&$77.47\pm_{0.99}$&$77.64\pm_{3.23}$&$46.5\pm_{0.16}$&$16.03\pm_{0.08}$&$6.64\pm_{0.12}$&$18.73\pm_{0.51}$\\
&TSMixer&$698.27\pm_{6.79}$&$146.22\pm_{23.62}$&$130.91\pm_{21.4}$&$88.39\pm_{4.38}$&$28.49\pm_{1.37}$&$16.46\pm_{8.55}$&$79.02\pm_{69.45}$\\
&FreTS&$419.07\pm_{6.85}$&$108.63\pm_{1.75}$&$98.46\pm_{14.62}$&$62.17\pm_{5.14}$&$20.54\pm_{1.82}$&$8.47\pm_{0.54}$&$25.89\pm_{1.53}$\\
&FiLM&$277.66\pm_{34.84}$&$78.58\pm_{6.24}$&$69.56\pm_{6.12}$&$45.37\pm_{2.5}$&$15.26\pm_{0.87}$&$6.79\pm_{0.39}$&$19.86\pm_{0.98}$\\
&Koopa&$282.85\pm_{6.89}$&$78.75\pm_{1.42}$&$83.25\pm_{7.31}$&$48.75\pm_{1.55}$&$16.92\pm_{0.47}$&$6.93\pm_{0.13}$&$19.16\pm_{0.11}$\\
\midrule
\multirow{16}{*}{\rotatebox{90}{RMSE}}&LSTM&$2148.33\pm_{0.79}$&$534.61\pm_{0.67}$&$704.41\pm_{136.89}$&$466.21\pm_{60.24}$&$178.29\pm_{26.25}$&$58.54\pm_{5.62}$&$192.72\pm_{30.76}$\\
&SegRNN&$1130.19\pm_{0.81}$&$304.2\pm_{2.76}$&$387.1\pm_{3.14}$&$264.53\pm_{10.99}$&$110.27\pm_{3.44}$&$36.38\pm_{1.34}$&$120.84\pm_{7.95}$\\
&CHGH&$2153.77\pm_{0.87}$&$537.97\pm_{0.18}$&$735.74\pm_{97.15}$&$468.27\pm_{61.0}$&$179.17\pm_{26.85}$&$41.0\pm_{28.76}$&$203.78\pm_{87.33}$\\
&PreDyGAE&$1510.41\pm_{0.01}$&$403.02\pm_{0.06}$&$529.39\pm_{58.83}$&$388.54\pm_{43.46}$&$159.45\pm_{15.71}$&$36.48\pm_{2.19}$&$162.0\pm_{20.48}$\\
&Transformer&$2154.53\pm_{30.61}$&$562.07\pm_{12.46}$&$717.62\pm_{125.1}$&$474.06\pm_{63.8}$&$179.2\pm_{25.64}$&$61.55\pm_{5.56}$&$191.91\pm_{36.18}$\\
&Autoformer&$2566.07\pm_{102.19}$&$489.69\pm_{10.56}$&$621.06\pm_{80.33}$&$387.57\pm_{33.41}$&$155.94\pm_{0.51}$&$58.1\pm_{4.51}$&$187.41\pm_{19.68}$\\
&Informer&$2212.83\pm_{2.36}$&$559.86\pm_{11.51}$&$715.12\pm_{132.16}$&$467.05\pm_{58.69}$&$178.41\pm_{28.75}$&$57.53\pm_{7.65}$&$191.85\pm_{36.5}$\\
&Reformer&$2248.94\pm_{15.68}$&$558.75\pm_{0.8}$&$713.91\pm_{131.53}$&$478.6\pm_{70.65}$&$179.85\pm_{27.33}$&$59.48\pm_{9.48}$&$195.6\pm_{31.87}$\\
&FEDformer&$2551.16\pm_{77.11}$&$501.84\pm_{4.46}$&$621.78\pm_{81.74}$&$363.84\pm_{35.37}$&$149.67\pm_{19.37}$&$57.89\pm_{2.21}$&$166.63\pm_{10.36}$\\
&NStransformer&$1337.16\pm_{47.86}$&$373.7\pm_{20.75}$&$473.97\pm_{10.71}$&$334.17\pm_{14.36}$&$128.79\pm_{8.11}$&$39.55\pm_{1.82}$&$176.86\pm_{17.5}$\\
&PatchTST&$1328.54\pm_{49.25}$&$347.07\pm_{11.24}$&$499.82\pm_{22.78}$&$321.16\pm_{15.89}$&$125.31\pm_{13.05}$&$39.74\pm_{1.04}$&$144.13\pm_{14.98}$\\
&DLinear&$1341.55\pm_{32.44}$&$350.8\pm_{6.91}$&$506.06\pm_{18.99}$&$325.15\pm_{16.25}$&$130.09\pm_{10.18}$&$38.99\pm_{1.16}$&$144.53\pm_{18.45}$\\
&TSMixer&$3040.93\pm_{86.91}$&$678.15\pm_{105.71}$&$816.57\pm_{184.07}$&$582.2\pm_{93.85}$&$211.98\pm_{22.73}$&$115.88\pm_{67.93}$&$436.64\pm_{301.19}$\\
&FreTS&$2105.11\pm_{31.62}$&$526.0\pm_{6.74}$&$679.7\pm_{130.16}$&$451.97\pm_{53.49}$&$173.34\pm_{22.83}$&$57.76\pm_{4.47}$&$188.29\pm_{25.82}$\\
&FiLM&$1298.06\pm_{265.59}$&$341.89\pm_{60.42}$&$426.14\pm_{19.46}$&$276.0\pm_{13.7}$&$111.46\pm_{9.34}$&$34.32\pm_{1.86}$&$128.54\pm_{16.33}$\\
&Koopa&$1386.85\pm_{35.21}$&$360.64\pm_{11.92}$&$532.64\pm_{14.5}$&$338.26\pm_{12.88}$&$138.63\pm_{9.35}$&$41.39\pm_{0.09}$&$152.48\pm_{17.55}$\\
\midrule
\multirow{16}{*}{\rotatebox{90}{SMAPE(\%)}}&LSTM&$37.23\pm_{0.05}$&$45.67\pm_{0.02}$&$45.95\pm_{1.99}$&$50.34\pm_{1.28}$&$47.36\pm_{0.66}$&$42.47\pm_{0.4}$&$70.15\pm_{1.62}$\\
&SegRNN&$32.88\pm_{0.2}$&$43.35\pm_{0.39}$&$41.54\pm_{2.12}$&$46.46\pm_{3.08}$&$47.11\pm_{1.77}$&$44.29\pm_{4.45}$&$59.48\pm_{1.85}$\\
&CHGH&$37.21\pm_{0.01}$&$45.55\pm_{0.03}$&$46.29\pm_{1.26}$&$50.3\pm_{1.28}$&$46.22\pm_{0.72}$&$45.92\pm_{21.79}$&$77.34\pm_{37.44}$\\
&PreDyGAE&$32.0\pm_{0.02}$&$43.25\pm_{0.18}$&$49.33\pm_{2.76}$&$51.02\pm_{5.06}$&$50.59\pm_{1.35}$&$63.2\pm_{17.41}$&$70.72\pm_{0.42}$\\
&Transformer&$44.18\pm_{0.71}$&$53.31\pm_{0.19}$&$53.13\pm_{2.85}$&$56.88\pm_{3.45}$&$55.61\pm_{8.32}$&$55.47\pm_{7.12}$&$73.21\pm_{7.57}$\\
&Autoformer&$67.46\pm_{0.27}$&$68.4\pm_{0.03}$&$70.67\pm_{6.58}$&$71.92\pm_{3.02}$&$83.21\pm_{4.31}$&$83.71\pm_{4.79}$&$99.62\pm_{14.04}$\\
&Informer&$43.4\pm_{0.31}$&$53.28\pm_{0.16}$&$53.54\pm_{2.22}$&$56.42\pm_{4.14}$&$56.23\pm_{7.69}$&$54.33\pm_{8.61}$&$73.34\pm_{7.52}$\\
&Reformer&$43.1\pm_{0.01}$&$53.09\pm_{0.1}$&$53.42\pm_{2.4}$&$56.23\pm_{4.31}$&$55.84\pm_{8.2}$&$52.75\pm_{10.9}$&$76.27\pm_{3.57}$\\
&FEDformer&$68.1\pm_{0.29}$&$67.4\pm_{0.48}$&$70.57\pm_{7.06}$&$70.35\pm_{5.46}$&$76.14\pm_{3.56}$&$79.91\pm_{0.11}$&$95.19\pm_{7.37}$\\
&NStransformer&$32.85\pm_{0.14}$&$42.35\pm_{0.05}$&$44.17\pm_{1.08}$&$47.32\pm_{0.29}$&$44.23\pm_{1.08}$&$38.15\pm_{1.31}$&$103.89\pm_{60.87}$\\
&PatchTST&$31.96\pm_{2.0}$&$41.14\pm_{2.29}$&$42.63\pm_{1.59}$&$45.79\pm_{2.31}$&$43.83\pm_{2.11}$&$37.55\pm_{2.57}$&$58.04\pm_{2.54}$\\
&DLinear&$31.38\pm_{1.08}$&$40.56\pm_{1.37}$&$41.8\pm_{1.23}$&$44.83\pm_{0.38}$&$45.42\pm_{0.14}$&$41.61\pm_{0.42}$&$58.22\pm_{0.52}$\\
&TSMixer&$68.55\pm_{1.22}$&$67.06\pm_{11.47}$&$61.4\pm_{6.47}$&$75.25\pm_{0.75}$&$72.63\pm_{11.4}$&$83.29\pm_{31.13}$&$109.31\pm_{46.77}$\\
&FreTS&$37.47\pm_{0.7}$&$46.12\pm_{0.91}$&$45.36\pm_{2.66}$&$50.03\pm_{1.63}$&$46.82\pm_{1.3}$&$41.25\pm_{1.28}$&$70.0\pm_{1.47}$\\
&FiLM&$32.01\pm_{0.16}$&$41.26\pm_{0.46}$&$41.12\pm_{1.97}$&$45.56\pm_{1.21}$&$45.99\pm_{0.95}$&$40.91\pm_{0.79}$&$58.02\pm_{1.51}$\\
&Koopa&$31.01\pm_{0.02}$&$39.92\pm_{0.13}$&$42.95\pm_{3.2}$&$45.33\pm_{1.65}$&$41.88\pm_{0.65}$&$35.91\pm_{0.3}$&$57.19\pm_{1.63}$\\
\midrule
\multirow{16}{*}{\rotatebox{90}{RRMSE(\%)}}&LSTM&$53.94\pm_{0.16}$&$53.9\pm_{0.12}$&$73.77\pm_{14.35}$&$98.43\pm_{12.75}$&$93.46\pm_{13.83}$&$65.39\pm_{6.3}$&$227.07\pm_{36.31}$\\
&SegRNN&$21.23\pm_{0.03}$&$22.98\pm_{0.43}$&$32.33\pm_{4.99}$&$41.84\pm_{8.49}$&$44.18\pm_{9.25}$&$34.49\pm_{6.31}$&$73.51\pm_{14.73}$\\
&CHGH&$54.21\pm_{0.03}$&$54.5\pm_{0.02}$&$77.63\pm_{10.33}$&$99.15\pm_{13.03}$&$93.96\pm_{14.27}$&$53.56\pm_{32.26}$&$105.96\pm_{104.02}$\\
&PreDyGAE&$30.72\pm_{0.01}$&$33.15\pm_{0.01}$&$46.59\pm_{9.64}$&$55.32\pm_{12.77}$&$57.56\pm_{12.47}$&$45.53\pm_{6.19}$&$91.47\pm_{23.2}$\\
&Transformer&$52.69\pm_{0.67}$&$56.3\pm_{1.62}$&$76.3\pm_{10.8}$&$99.08\pm_{12.9}$&$94.57\pm_{7.31}$&$67.56\pm_{5.78}$&$215.96\pm_{35.18}$\\
&Autoformer&$51.07\pm_{0.19}$&$39.44\pm_{1.34}$&$54.75\pm_{0.61}$&$65.4\pm_{1.15}$&$64.75\pm_{3.85}$&$54.46\pm_{8.57}$&$106.51\pm_{5.39}$\\
&Informer&$55.42\pm_{0.04}$&$56.22\pm_{2.17}$&$71.7\pm_{12.14}$&$98.81\pm_{8.31}$&$93.29\pm_{11.35}$&$64.2\pm_{6.07}$&$213.07\pm_{32.78}$\\
&Reformer&$56.91\pm_{0.64}$&$55.93\pm_{0.41}$&$76.53\pm_{13.64}$&$99.47\pm_{13.64}$&$95.12\pm_{10.02}$&$67.05\pm_{7.72}$&$221.45\pm_{36.6}$\\
&FEDformer&$49.35\pm_{2.5}$&$40.99\pm_{1.28}$&$56.8\pm_{2.22}$&$61.57\pm_{0.66}$&$63.27\pm_{4.31}$&$54.5\pm_{0.19}$&$105.93\pm_{4.82}$\\
&NStransformer&$27.47\pm_{1.23}$&$31.55\pm_{2.22}$&$43.9\pm_{3.25}$&$60.16\pm_{1.54}$&$56.5\pm_{1.0}$&$39.75\pm_{0.02}$&$116.26\pm_{2.18}$\\
&PatchTST&$27.53\pm_{1.1}$&$29.31\pm_{1.0}$&$47.82\pm_{3.34}$&$57.57\pm_{1.93}$&$55.03\pm_{2.9}$&$40.96\pm_{1.97}$&$114.76\pm_{7.12}$\\
&DLinear&$27.83\pm_{1.38}$&$29.66\pm_{1.38}$&$49.37\pm_{3.5}$&$59.59\pm_{1.4}$&$59.48\pm_{0.62}$&$40.74\pm_{0.89}$&$116.83\pm_{1.45}$\\
&TSMixer&$76.26\pm_{6.24}$&$69.83\pm_{15.63}$&$86.8\pm_{23.27}$&$119.17\pm_{17.64}$&$113.43\pm_{16.08}$&$77.61\pm_{14.65}$&$141.79\pm_{36.12}$\\
&FreTS&$52.23\pm_{1.04}$&$52.51\pm_{0.92}$&$69.87\pm_{13.95}$&$93.32\pm_{11.09}$&$88.52\pm_{11.95}$&$63.64\pm_{5.3}$&$215.64\pm_{25.29}$\\
&FiLM&$26.91\pm_{7.65}$&$28.84\pm_{7.48}$&$36.38\pm_{1.32}$&$43.93\pm_{0.54}$&$44.44\pm_{1.26}$&$31.45\pm_{0.6}$&$83.15\pm_{2.17}$\\
&Koopa&$28.98\pm_{1.27}$&$30.77\pm_{1.6}$&$53.08\pm_{5.03}$&$63.24\pm_{2.77}$&$65.24\pm_{0.87}$&$44.14\pm_{3.04}$&$129.36\pm_{2.2}$\\
\bottomrule
\end{tabular}
}
    \label{tab_main:break}
\end{table}

Table \ref{tab_main:break} presents the results of repeated experiments on forecasting skill demand sequences that have undergone structural breaks. Initially, the overall errors are quite pronounced, underscoring the challenge of accurately predicting these skills. Moreover, FiLM performs well on most metrics, which further verifies its robustness.

\subsection{Job Skill Demand Forecasting for Low-Frequency Skills}
In multigranular skill demand sequences, a significant number of skills remain inactive or in low frequency over extended periods. These skills might continue to have low demand in the future (indicating low importance), or they might suddenly gain interest from certain professions or companies, leading to rapid growth. In this study, we define low-frequency skills as those that appear fewer than twice in the time slices of the training set. Predicting the demand for these skills is challenging because their data points are predominantly zero during training, resulting in a lack of effective observational data. Therefore, we specifically present the demand prediction results of the existing benchmark models for these low-frequency skills.

\paragraph{Results.}
We continued to test the demand prediction effect on low-frequency skills using the benchmark models described in the main text, and the results are shown in Table \ref{tab:low-freq}. From this, we can draw the following conclusions: Firstly, there is a significant increase in the error on the RRMSE metric, indicating that low-demand skills are difficult to predict accurately. Secondly, Koopa has the best predictive performance in this scenario. We also found that the performance of SegRNN significantly decreases, suggesting that SegRNN's segment learning approach is not suitable for predicting low-frequency skill demands due to a lack of effective observational data, rendering the learning segments meaningless.

\begin{table}[ht!]
    \centering
    \caption{Performance comparisons on skill demand series with low-frequency.}
\resizebox{\textwidth}{!}{
\begin{tabular}{l|llllllll}
\toprule
&Model&Market&Region&L1-O&L2-O&R\&L1-O&R\&L2-O&Company\\
\midrule
\multirow{16}{*}{\rotatebox{90}{MAE}}&LSTM&$33.69\pm_{0.44}$&$26.88\pm_{0.05}$&$16.49\pm_{0.01}$&$12.11\pm_{0.04}$&$12.28\pm_{0.02}$&$8.92\pm_{0.1}$&$12.6\pm_{0.02}$\\
&SegRNN&$51.5\pm_{0.57}$&$46.12\pm_{0.28}$&$25.65\pm_{0.39}$&$18.03\pm_{0.03}$&$22.08\pm_{1.92}$&$19.1\pm_{3.22}$&$19.37\pm_{0.88}$\\
&CHGH&$32.37\pm_{0.0}$&$24.99\pm_{0.02}$&$14.2\pm_{0.04}$&$9.32\pm_{0.01}$&$9.45\pm_{0.0}$&$67.89\pm_{10.03}$&$77.01\pm_{5.58}$\\
&PreDyGAE&$113.67\pm_{1.73}$&$157.17\pm_{18.68}$&$343.32\pm_{0.15}$&$64.46\pm_{13.49}$&$24.16\pm_{0.13}$&$60.41\pm_{55.15}$&$68.89\pm_{62.89}$\\
&Transformer&$54.84\pm_{1.99}$&$52.95\pm_{0.09}$&$46.1\pm_{0.13}$&$45.47\pm_{0.07}$&$45.64\pm_{0.02}$&$45.1\pm_{0.01}$&$45.43\pm_{0.0}$\\
&Autoformer&$132.98\pm_{11.42}$&$106.13\pm_{1.65}$&$110.06\pm_{0.19}$&$97.66\pm_{1.23}$&$98.28\pm_{0.46}$&$95.26\pm_{1.1}$&$90.23\pm_{0.54}$\\
&Informer&$54.42\pm_{1.83}$&$52.76\pm_{0.13}$&$46.1\pm_{0.0}$&$45.52\pm_{0.01}$&$45.69\pm_{0.02}$&$45.07\pm_{0.01}$&$45.47\pm_{0.0}$\\
&Reformer&$52.72\pm_{0.3}$&$53.25\pm_{0.15}$&$46.19\pm_{0.07}$&$45.58\pm_{0.02}$&$45.65\pm_{0.01}$&$45.08\pm_{0.03}$&$45.49\pm_{0.03}$\\
&FEDformer&$133.07\pm_{6.09}$&$101.47\pm_{1.47}$&$109.51\pm_{1.3}$&$96.43\pm_{1.0}$&$95.83\pm_{1.48}$&$94.18\pm_{0.48}$&$90.71\pm_{0.47}$\\
&NStransformer&$57.96\pm_{1.83}$&$34.24\pm_{0.3}$&$17.99\pm_{0.37}$&$10.84\pm_{0.04}$&$11.46\pm_{0.02}$&$6.45\pm_{0.02}$&$11.9\pm_{0.0}$\\
&PatchTST&$47.25\pm_{6.07}$&$33.21\pm_{3.3}$&$17.31\pm_{1.85}$&$10.39\pm_{1.03}$&$10.91\pm_{1.1}$&$6.1\pm_{0.67}$&$11.4\pm_{1.23}$\\
&DLinear&$51.7\pm_{7.31}$&$42.08\pm_{8.83}$&$29.64\pm_{9.87}$&$25.18\pm_{10.9}$&$25.6\pm_{10.9}$&$22.53\pm_{11.72}$&$25.79\pm_{10.77}$\\
&TSMixer&$102.23\pm_{14.26}$&$64.77\pm_{10.8}$&$51.89\pm_{9.94}$&$52.61\pm_{1.14}$&$40.06\pm_{7.0}$&$34.02\pm_{2.2}$&$41.32\pm_{2.03}$\\
&FreTS&$38.55\pm_{0.83}$&$28.89\pm_{0.96}$&$16.6\pm_{0.58}$&$11.19\pm_{0.47}$&$11.46\pm_{0.51}$&$7.44\pm_{0.47}$&$11.94\pm_{0.48}$\\
&FiLM&$51.42\pm_{20.47}$&$34.97\pm_{10.99}$&$18.23\pm_{5.69}$&$10.9\pm_{3.11}$&$11.47\pm_{3.38}$&$6.43\pm_{1.96}$&$11.95\pm_{3.58}$\\
&Koopa&$43.3\pm_{4.56}$&$29.66\pm_{2.37}$&$15.52\pm_{1.38}$&$9.49\pm_{0.73}$&$9.95\pm_{0.82}$&$5.53\pm_{0.48}$&$10.35\pm_{0.86}$\\
\midrule
\multirow{16}{*}{\rotatebox{90}{RMSE}}&LSTM&$99.83\pm_{0.77}$&$306.02\pm_{0.33}$&$168.19\pm_{0.01}$&$140.57\pm_{0.0}$&$247.24\pm_{0.02}$&$125.6\pm_{0.01}$&$272.77\pm_{0.04}$\\
&SegRNN&$141.21\pm_{0.31}$&$452.49\pm_{0.09}$&$233.37\pm_{0.06}$&$146.34\pm_{0.22}$&$216.77\pm_{0.01}$&$117.81\pm_{0.63}$&$227.59\pm_{0.0}$\\
&CHGH&$100.44\pm_{0.02}$&$306.28\pm_{0.01}$&$168.23\pm_{0.01}$&$140.54\pm_{0.0}$&$247.21\pm_{0.01}$&$134.20\pm_{0.0}$&$208.01\pm_{17.09}$\\
&PreDyGAE&$169.76\pm_{1.46}$&$542.42\pm_{24.84}$&$771.72\pm_{50.15}$&$195.02\pm_{16.23}$&$218.94\pm_{0.65}$&$119.76\pm_{109.33}$&$188.17\pm_{171.77}$\\
&Transformer&$112.3\pm_{5.16}$&$311.19\pm_{1.43}$&$176.74\pm_{0.52}$&$150.63\pm_{0.53}$&$252.9\pm_{0.1}$&$137.39\pm_{0.13}$&$278.45\pm_{0.13}$\\
&Autoformer&$249.03\pm_{36.15}$&$364.32\pm_{9.52}$&$231.67\pm_{3.95}$&$184.29\pm_{1.11}$&$266.1\pm_{0.64}$&$172.38\pm_{1.84}$&$281.53\pm_{0.17}$\\
&Informer&$111.59\pm_{6.84}$&$312.15\pm_{0.35}$&$176.18\pm_{0.19}$&$150.46\pm_{0.16}$&$252.84\pm_{0.01}$&$137.07\pm_{0.12}$&$279.56\pm_{0.36}$\\
&Reformer&$107.11\pm_{3.02}$&$314.04\pm_{1.77}$&$176.73\pm_{0.57}$&$150.62\pm_{0.23}$&$252.85\pm_{0.14}$&$137.02\pm_{0.12}$&$279.79\pm_{0.61}$\\
&FEDformer&$233.86\pm_{32.3}$&$346.25\pm_{1.54}$&$227.64\pm_{2.23}$&$182.54\pm_{0.06}$&$264.12\pm_{0.64}$&$171.43\pm_{0.76}$&$280.57\pm_{1.13}$\\
&NStransformer&$159.4\pm_{10.66}$&$329.86\pm_{4.31}$&$182.83\pm_{8.95}$&$136.01\pm_{2.69}$&$249.63\pm_{0.44}$&$129.74\pm_{5.67}$&$255.48\pm_{4.76}$\\
&PatchTST&$126.71\pm_{17.24}$&$339.92\pm_{16.66}$&$182.49\pm_{8.84}$&$134.57\pm_{2.89}$&$233.14\pm_{1.45}$&$120.94\pm_{1.7}$&$253.85\pm_{3.35}$\\
&DLinear&$109.06\pm_{5.5}$&$320.09\pm_{2.51}$&$173.51\pm_{2.67}$&$133.82\pm_{2.37}$&$235.23\pm_{1.35}$&$123.01\pm_{2.84}$&$254.7\pm_{1.91}$\\
&TSMixer&$186.01\pm_{18.76}$&$359.57\pm_{28.14}$&$199.07\pm_{10.44}$&$160.52\pm_{5.4}$&$261.65\pm_{6.67}$&$139.7\pm_{3.15}$&$296.15\pm_{1.83}$\\
&FreTS&$104.43\pm_{1.38}$&$310.94\pm_{2.43}$&$170.36\pm_{1.47}$&$139.96\pm_{0.74}$&$239.77\pm_{4.49}$&$122.58\pm_{2.07}$&$266.82\pm_{3.88}$\\
&FiLM&$129.11\pm_{32.28}$&$343.63\pm_{34.15}$&$185.3\pm_{16.68}$&$135.8\pm_{1.72}$&$235.13\pm_{10.73}$&$122.84\pm_{2.06}$&$256.54\pm_{7.91}$\\
&Koopa&$111.96\pm_{8.52}$&$317.8\pm_{3.82}$&$171.28\pm_{3.23}$&$134.11\pm_{0.67}$&$242.79\pm_{2.91}$&$124.64\pm_{0.36}$&$260.87\pm_{0.46}$\\
\midrule
\multirow{16}{*}{\rotatebox{90}{SMAPE(\%)}}&LSTM&$26.1\pm_{0.39}$&$19.43\pm_{0.17}$&$15.18\pm_{0.01}$&$12.88\pm_{0.07}$&$13.01\pm_{0.02}$&$11.46\pm_{0.17}$&$13.01\pm_{0.03}$\\
&SegRNN&$33.71\pm_{0.71}$&$29.55\pm_{0.4}$&$22.04\pm_{0.56}$&$18.44\pm_{0.11}$&$24.16\pm_{2.97}$&$24.8\pm_{4.63}$&$19.83\pm_{1.32}$\\
&CHGH&$23.61\pm_{0.0}$&$15.99\pm_{0.04}$&$11.08\pm_{0.09}$&$7.8\pm_{0.03}$&$7.86\pm_{0.01}$&$9.61\pm_{1.08}$&$8.07\pm_{2.03}$\\
&PreDyGAE&$83.59\pm_{0.72}$&$82.33\pm_{4.38}$&$114.25\pm_{1.42}$&$48.0\pm_{7.33}$&$23.97\pm_{0.1}$&$47.34\pm_{43.22}$&$48.31\pm_{44.1}$\\
&Transformer&$49.05\pm_{1.26}$&$48.23\pm_{0.0}$&$48.68\pm_{0.14}$&$50.78\pm_{0.03}$&$50.92\pm_{0.01}$&$52.6\pm_{0.01}$&$50.28\pm_{0.0}$\\
&Autoformer&$80.95\pm_{1.5}$&$76.73\pm_{0.44}$&$81.69\pm_{0.03}$&$80.14\pm_{0.55}$&$80.43\pm_{0.21}$&$81.25\pm_{0.49}$&$76.73\pm_{0.25}$\\
&Informer&$48.87\pm_{0.78}$&$48.07\pm_{0.25}$&$48.69\pm_{0.03}$&$50.83\pm_{0.01}$&$50.96\pm_{0.02}$&$52.57\pm_{0.01}$&$50.29\pm_{0.0}$\\
&Reformer&$48.32\pm_{1.06}$&$48.33\pm_{0.02}$&$48.76\pm_{0.08}$&$50.87\pm_{0.03}$&$50.93\pm_{0.0}$&$52.59\pm_{0.02}$&$50.28\pm_{0.02}$\\
&FEDformer&$82.56\pm_{1.23}$&$75.32\pm_{0.52}$&$81.88\pm_{0.57}$&$79.7\pm_{0.39}$&$79.55\pm_{0.62}$&$80.83\pm_{0.21}$&$76.96\pm_{0.21}$\\
&NStransformer&$32.62\pm_{0.69}$&$18.84\pm_{0.31}$&$11.59\pm_{0.02}$&$7.3\pm_{0.01}$&$7.81\pm_{0.01}$&$4.76\pm_{0.0}$&$7.76\pm_{0.0}$\\
&PatchTST&$29.48\pm_{1.53}$&$18.23\pm_{1.46}$&$11.15\pm_{1.04}$&$7.01\pm_{0.71}$&$7.51\pm_{0.76}$&$4.56\pm_{0.52}$&$7.46\pm_{0.8}$\\
&DLinear&$42.19\pm_{8.16}$&$35.1\pm_{10.8}$&$31.64\pm_{12.55}$&$30.55\pm_{14.04}$&$30.93\pm_{14.03}$&$30.35\pm_{15.14}$&$30.52\pm_{13.82}$\\
&TSMixer&$68.79\pm_{5.48}$&$51.84\pm_{6.45}$&$48.47\pm_{6.43}$&$54.61\pm_{0.92}$&$45.87\pm_{6.93}$&$43.34\pm_{2.84}$&$45.68\pm_{1.95}$\\
&FreTS&$29.63\pm_{0.33}$&$19.71\pm_{0.62}$&$14.01\pm_{0.56}$&$10.6\pm_{0.64}$&$10.86\pm_{0.67}$&$8.45\pm_{0.75}$&$11.09\pm_{0.61}$\\
&FiLM&$30.74\pm_{10.37}$&$18.56\pm_{6.37}$&$11.42\pm_{4.06}$&$7.19\pm_{2.55}$&$7.7\pm_{2.79}$&$4.69\pm_{1.69}$&$7.64\pm_{2.84}$\\
&Koopa&$28.27\pm_{2.34}$&$16.67\pm_{1.38}$&$10.15\pm_{0.9}$&$6.35\pm_{0.57}$&$6.79\pm_{0.63}$&$4.09\pm_{0.39}$&$6.73\pm_{0.63}$\\
\midrule
\multirow{16}{*}{\rotatebox{90}{RRMSE(\%)}}&LSTM&$432.63\pm_{21.64}$&$1364.7\pm_{202.93}$&$1057.49\pm_{28.76}$&$1468.93\pm_{15.86}$&$2581.13\pm_{33.65}$&$1485.07\pm_{28.13}$&$1828.92\pm_{33.42}$\\
&SegRNN&$114.7\pm_{0.21}$&$125.97\pm_{0.16}$&$126.24\pm_{0.18}$&$146.81\pm_{0.73}$&$130.23\pm_{0.45}$&$130.24\pm_{0.87}$&$133.03\pm_{0.54}$\\
&CHGH&$452.99\pm_{0.65}$&$1662.7\pm_{2.11}$&$1394.35\pm_{4.28}$&$1824.0\pm_{1.1}$&$3436.23\pm_{14.46}$&$289.09\pm_{14.07}$&$406.07\pm_{8.08}$\\
&PreDyGAE&$104.17\pm_{0.99}$&$216.23\pm_{2.0}$&$201.73\pm_{0.3}$&$220.44\pm_{4.62}$&$229.42\pm_{0.22}$&$165.21\pm_{59.52}$&$272.3\pm_{66.0}$\\
&Transformer&$172.07\pm_{30.78}$&$513.13\pm_{29.31}$&$320.11\pm_{12.98}$&$269.97\pm_{3.45}$&$468.03\pm_{0.41}$&$242.99\pm_{4.1}$&$466.5\pm_{1.48}$\\
&Autoformer&$89.47\pm_{0.4}$&$161.16\pm_{10.32}$&$136.15\pm_{6.04}$&$134.91\pm_{1.17}$&$195.47\pm_{1.88}$&$134.43\pm_{1.31}$&$191.17\pm_{2.0}$\\
&Informer&$193.88\pm_{30.22}$&$538.02\pm_{31.16}$&$316.88\pm_{4.27}$&$275.64\pm_{0.03}$&$470.15\pm_{2.27}$&$246.67\pm_{0.09}$&$462.58\pm_{1.78}$\\
&Reformer&$200.52\pm_{32.93}$&$536.21\pm_{13.15}$&$324.86\pm_{11.04}$&$276.66\pm_{0.14}$&$470.18\pm_{0.3}$&$247.32\pm_{0.89}$&$460.86\pm_{4.08}$\\
&FEDformer&$96.72\pm_{6.76}$&$176.64\pm_{0.39}$&$136.62\pm_{1.69}$&$136.93\pm_{0.31}$&$199.25\pm_{1.04}$&$134.8\pm_{0.73}$&$189.29\pm_{1.49}$\\
&NStransformer&$90.14\pm_{7.16}$&$219.18\pm_{39.19}$&$193.48\pm_{36.91}$&$218.73\pm_{0.68}$&$406.29\pm_{7.04}$&$272.29\pm_{31.63}$&$259.15\pm_{8.34}$\\
&PatchTST&$95.85\pm_{5.0}$&$210.2\pm_{36.63}$&$199.12\pm_{34.35}$&$266.55\pm_{38.3}$&$391.36\pm_{46.33}$&$298.48\pm_{41.01}$&$278.87\pm_{33.7}$\\
&DLinear&$90.72\pm_{2.96}$&$241.41\pm_{14.46}$&$219.68\pm_{17.61}$&$281.05\pm_{37.87}$&$431.86\pm_{48.87}$&$296.1\pm_{49.89}$&$300.39\pm_{25.41}$\\
&TSMixer&$125.99\pm_{9.23}$&$200.39\pm_{22.34}$&$211.03\pm_{36.38}$&$192.85\pm_{35.49}$&$418.03\pm_{16.7}$&$275.29\pm_{1.9}$&$293.38\pm_{24.47}$\\
&FreTS&$222.55\pm_{30.4}$&$479.45\pm_{20.87}$&$462.19\pm_{33.86}$&$720.43\pm_{51.2}$&$962.73\pm_{47.52}$&$758.24\pm_{41.67}$&$809.76\pm_{76.22}$\\
&FiLM&$108.05\pm_{34.05}$&$444.72\pm_{401.85}$&$360.89\pm_{298.05}$&$402.84\pm_{297.56}$&$652.7\pm_{546.99}$&$417.27\pm_{297.85}$&$364.47\pm_{235.25}$\\
&Koopa&$93.83\pm_{0.11}$&$287.42\pm_{57.85}$&$264.45\pm_{56.56}$&$345.86\pm_{79.42}$&$582.52\pm_{154.25}$&$368.66\pm_{64.94}$&$345.87\pm_{66.81}$\\
\bottomrule
\end{tabular}
}
    \label{tab:low-freq}
\end{table}

\begin{table}[ht!]
    \centering
    \caption{Performance comparisons on skill demand series with GNN-based methods.}
\resizebox{\textwidth}{!}{
\begin{tabular}{l|l|lllllll}
\toprule
&{Model}&Market&Region&L1-O&L2-O&R\&L1-O&R\&L2-O&Company\\
\midrule
\multirow{7}{*}{\rotatebox{90}{MAE}}&EvolveGCNH&$1053.18_{\pm804.38}$&$30.07_{\pm22.97}$&$16.51_{\pm12.61}$&$5.43_{\pm4.15}$&$2.92_{\pm2.23}$&$1.11_{\pm0.85}$&$0.96_{\pm0.73}$\\
&EvolveGCNO&$151.13_{\pm115.43}$&$27.01_{\pm20.63}$&$15.4_{\pm11.76}$&$5.04_{\pm3.85}$&$2.95_{\pm2.25}$&$1.21_{\pm0.92}$&$0.89_{\pm0.68}$\\
&GConvGRU&$570.31_{\pm435.58}$&$92.96_{\pm71.0}$&$46.43_{\pm35.46}$&$11.45_{\pm8.75}$&$5.47_{\pm4.18}$&$1.4_{\pm1.07}$&$1.04_{\pm0.79}$\\
&TGCN&$741.05_{\pm565.99}$&$96.43_{\pm73.65}$&$55.41_{\pm42.32}$&$13.21_{\pm10.09}$&$6.28_{\pm4.8}$&$1.63_{\pm1.24}$&$1.11_{\pm0.85}$\\
&GCLSTM&$729.48_{\pm557.15}$&$57.13_{\pm43.63}$&$25.84_{\pm19.74}$&$11.49_{\pm8.78}$&$5.67_{\pm4.33}$&$1.61_{\pm1.23}$&$1.07_{\pm0.82}$\\
&GConvLSTM&$741.45_{\pm566.29}$&$93.61_{\pm71.5}$&$46.57_{\pm35.57}$&$11.19_{\pm8.55}$&$5.57_{\pm4.25}$&$1.49_{\pm1.14}$&$1.09_{\pm0.83}$\\
&DyGrEncoder&$732.53_{\pm559.48}$&$92.21_{\pm70.43}$&$46.96_{\pm35.87}$&$12.26_{\pm9.36}$&$5.57_{\pm4.26}$&$1.47_{\pm1.13}$&$1.07_{\pm0.82}$\\
\midrule
\multirow{7}{*}{\rotatebox{90}{RMSE}}&EvolveGCNH&$4998.7_{\pm3817.82}$&$166.34_{\pm127.04}$&$178.55_{\pm136.37}$&$76.14_{\pm58.15}$&$35.12_{\pm26.82}$&$16.97_{\pm12.96}$&$19.33_{\pm14.77}$\\
&EvolveGCNO&$709.19_{\pm541.65}$&$143.72_{\pm109.77}$&$164.16_{\pm125.38}$&$71.57_{\pm54.67}$&$33.21_{\pm25.37}$&$15.04_{\pm11.49}$&$17.93_{\pm13.7}$\\
&GConvGRU&$2968.75_{\pm2267.42}$&$579.66_{\pm442.72}$&$442.53_{\pm337.99}$&$170.09_{\pm129.91}$&$81.95_{\pm62.59}$&$31.75_{\pm24.25}$&$30.65_{\pm23.41}$\\
&TGCN&$3163.46_{\pm2416.13}$&$581.49_{\pm444.12}$&$450.79_{\pm344.29}$&$172.13_{\pm131.47}$&$83.33_{\pm63.64}$&$32.19_{\pm24.59}$&$30.77_{\pm23.5}$\\
&GCLSTM&$3159.61_{\pm2413.19}$&$497.97_{\pm380.33}$&$371.13_{\pm283.45}$&$170.23_{\pm130.01}$&$82.66_{\pm63.13}$&$32.25_{\pm24.63}$&$30.81_{\pm23.53}$\\
&GConvLSTM&$3166.41_{\pm2418.38}$&$581.26_{\pm443.94}$&$442.72_{\pm338.13}$&$169.43_{\pm129.4}$&$82.38_{\pm62.92}$&$32.29_{\pm24.66}$&$30.88_{\pm23.58}$\\
&DyGrEncoder&$3161.61_{\pm2414.72}$&$579.78_{\pm442.81}$&$443.49_{\pm338.72}$&$171.91_{\pm131.3}$&$82.27_{\pm62.83}$&$32.02_{\pm24.46}$&$30.76_{\pm23.49}$\\
\midrule
\multirow{7}{*}{\rotatebox{90}{SMAPE(\%)}}&EvolveGCNH&$76.69_{\pm58.58}$&$41.67_{\pm31.83}$&$39.77_{\pm30.38}$&$33.11_{\pm25.29}$&$28.99_{\pm22.14}$&$18.79_{\pm14.35}$&$28.36_{\pm21.66}$\\
&EvolveGCNO&$45.32_{\pm34.61}$&$56.75_{\pm43.35}$&$46.34_{\pm35.39}$&$35.09_{\pm26.8}$&$29.33_{\pm22.4}$&$33.19_{\pm25.35}$&$25.99_{\pm19.85}$\\
&GConvGRU&$63.93_{\pm48.82}$&$62.41_{\pm47.67}$&$38.65_{\pm29.52}$&$31.09_{\pm23.74}$&$23.67_{\pm18.08}$&$12.93_{\pm9.88}$&$20.5_{\pm15.66}$\\
&TGCN&$71.13_{\pm54.32}$&$51.37_{\pm39.23}$&$56.23_{\pm42.95}$&$37.21_{\pm28.42}$&$25.83_{\pm19.73}$&$21.74_{\pm16.6}$&$23.35_{\pm17.83}$\\
&GCLSTM&$60.49_{\pm46.2}$&$38.68_{\pm29.54}$&$43.6_{\pm33.3}$&$28.15_{\pm21.5}$&$24.34_{\pm18.59}$&$18.57_{\pm14.18}$&$21.74_{\pm16.6}$\\
&GConvLSTM&$64.79_{\pm49.49}$&$44.77_{\pm34.2}$&$35.9_{\pm27.42}$&$26.21_{\pm20.02}$&$21.67_{\pm16.55}$&$13.95_{\pm10.66}$&$23.05_{\pm17.61}$\\
&DyGrEncoder&$61.55_{\pm47.01}$&$44.74_{\pm34.17}$&$38.12_{\pm29.11}$&$26.94_{\pm20.58}$&$26.49_{\pm20.23}$&$18.42_{\pm14.07}$&$23.75_{\pm18.14}$\\
\midrule
\multirow{7}{*}{\rotatebox{90}{RRMSE(\%)}}&EvolveGCNH&$68.68_{\pm52.46}$&$19.57_{\pm14.95}$&$32.99_{\pm25.2}$&$36.53_{\pm27.9}$&$32.1_{\pm24.52}$&$41.01_{\pm31.32}$&$65.93_{\pm50.35}$\\
&EvolveGCNO&$16.21_{\pm12.38}$&$17.85_{\pm13.63}$&$28.93_{\pm22.09}$&$34.6_{\pm26.43}$&$30.4_{\pm23.22}$&$32.27_{\pm24.64}$&$58.07_{\pm44.35}$\\
&GConvGRU&$418.84_{\pm319.89}$&$989.03_{\pm755.38}$&$1111.69_{\pm849.06}$&$701.09_{\pm535.46}$&$414.17_{\pm316.33}$&$307.07_{\pm234.53}$&$385.69_{\pm294.58}$\\
&TGCN&$3478.62_{\pm2656.84}$&$1037.13_{\pm792.12}$&$112164.83_{\pm85667.3}$&$948.04_{\pm724.08}$&$501.07_{\pm382.7}$&$338.75_{\pm258.73}$&$404.5_{\pm308.94}$\\
&GCLSTM&$3083.62_{\pm2355.15}$&$176.53_{\pm134.83}$&$159.96_{\pm122.17}$&$721.9_{\pm551.36}$&$467.49_{\pm357.05}$&$334.26_{\pm255.3}$&$425.46_{\pm324.95}$\\
&GConvLSTM&$4479.9_{\pm3421.58}$&$1123.32_{\pm857.95}$&$1101.0_{\pm840.9}$&$640.31_{\pm489.05}$&$449.12_{\pm343.02}$&$382.23_{\pm291.94}$&$451.71_{\pm345.0}$\\
&DyGrEncoder&$3386.49_{\pm2586.47}$&$1010.93_{\pm772.11}$&$1247.51_{\pm952.8}$&$1004.62_{\pm767.29}$&$450.57_{\pm344.13}$&$331.31_{\pm253.04}$&$418.67_{\pm319.77}$\\
\bottomrule
\end{tabular}
}
    \label{tab:graph_overall}
\end{table}
\begin{table}[ht!]
    \centering
    \caption{Performance comparisons on skill demand series with structural breaks with GNN-based methods.}
\resizebox{\textwidth}{!}{
\begin{tabular}{l|l|lllllll}
\toprule
&{Model}&Market&Region&L1-O&L2-O&R\&L1-O&R\&L2-O&Company\\
\midrule
\multirow{7}{*}{\rotatebox{90}{MAE}}&EvolveGCNH&$1385.2_{\pm1057.96}$&$61.79_{\pm47.19}$&$58.05_{\pm44.33}$&$36.04_{\pm27.53}$&$10.69_{\pm8.17}$&$5.56_{\pm4.25}$&$13.25_{\pm10.12}$\\
&EvolveGCNO&$195.87_{\pm149.6}$&$53.86_{\pm41.14}$&$52.46_{\pm40.07}$&$32.76_{\pm25.02}$&$10.85_{\pm8.28}$&$5.21_{\pm3.98}$&$12.22_{\pm9.33}$\\
&GConvGRU&$749.83_{\pm572.69}$&$204.83_{\pm156.44}$&$189.48_{\pm144.72}$&$92.29_{\pm70.49}$&$26.94_{\pm20.58}$&$9.23_{\pm7.05}$&$17.26_{\pm13.18}$\\
&TGCN&$990.21_{\pm756.28}$&$210.7_{\pm160.92}$&$222.48_{\pm169.92}$&$100.93_{\pm77.09}$&$29.27_{\pm22.36}$&$9.93_{\pm7.58}$&$17.87_{\pm13.65}$\\
&GCLSTM&$977.98_{\pm746.94}$&$128.33_{\pm98.02}$&$102.44_{\pm78.24}$&$92.71_{\pm70.81}$&$27.94_{\pm21.34}$&$10.25_{\pm7.83}$&$17.81_{\pm13.61}$\\
&GConvLSTM&$993.55_{\pm758.84}$&$207.86_{\pm158.76}$&$190.31_{\pm145.35}$&$90.33_{\pm68.99}$&$27.51_{\pm21.01}$&$9.93_{\pm7.58}$&$18.0_{\pm13.75}$\\
&DyGrEncoder&$982.05_{\pm750.06}$&$204.89_{\pm156.49}$&$191.91_{\pm146.57}$&$98.78_{\pm75.44}$&$27.41_{\pm20.93}$&$9.6_{\pm7.33}$&$17.51_{\pm13.38}$\\
\midrule
\multirow{7}{*}{\rotatebox{90}{RMSE}}&EvolveGCNH&$5799.81_{\pm4429.68}$&$261.83_{\pm199.97}$&$371.99_{\pm284.11}$&$227.88_{\pm174.05}$&$87.98_{\pm67.2}$&$29.87_{\pm22.82}$&$100.77_{\pm76.96}$\\
&EvolveGCNO&$863.11_{\pm659.21}$&$226.98_{\pm173.36}$&$341.34_{\pm260.7}$&$213.99_{\pm163.44}$&$82.25_{\pm62.82}$&$25.03_{\pm19.11}$&$93.36_{\pm71.3}$\\
&GConvGRU&$3463.23_{\pm2645.08}$&$890.59_{\pm680.2}$&$922.56_{\pm704.62}$&$508.61_{\pm388.46}$&$198.01_{\pm151.24}$&$69.96_{\pm53.43}$&$158.49_{\pm121.05}$\\
&TGCN&$3688.76_{\pm2817.34}$&$893.17_{\pm682.17}$&$938.93_{\pm717.12}$&$514.25_{\pm392.77}$&$201.03_{\pm153.54}$&$71.44_{\pm54.56}$&$159.07_{\pm121.49}$\\
&GCLSTM&$3684.35_{\pm2813.97}$&$768.16_{\pm586.69}$&$776.42_{\pm593.0}$&$508.99_{\pm388.74}$&$199.63_{\pm152.47}$&$71.71_{\pm54.77}$&$159.26_{\pm121.64}$\\
&GConvLSTM&$3692.17_{\pm2819.94}$&$892.93_{\pm681.98}$&$922.93_{\pm704.9}$&$506.71_{\pm387.0}$&$198.99_{\pm151.98}$&$71.93_{\pm54.94}$&$159.59_{\pm121.89}$\\
&DyGrEncoder&$3686.65_{\pm2815.72}$&$890.76_{\pm680.33}$&$924.49_{\pm706.09}$&$513.77_{\pm392.4}$&$198.73_{\pm151.79}$&$70.95_{\pm54.19}$&$159.01_{\pm121.44}$\\
\midrule
\multirow{7}{*}{\rotatebox{90}{SMAPE(\%)}}&EvolveGCNH&$77.85_{\pm59.46}$&$31.5_{\pm24.06}$&$30.67_{\pm23.43}$&$33.05_{\pm25.24}$&$33.43_{\pm25.54}$&$29.61_{\pm22.61}$&$38.66_{\pm29.53}$\\
&EvolveGCNO&$31.65_{\pm24.17}$&$33.88_{\pm25.88}$&$30.29_{\pm23.14}$&$31.57_{\pm24.11}$&$34.17_{\pm26.1}$&$37.43_{\pm28.59}$&$37.69_{\pm28.78}$\\
&GConvGRU&$49.4_{\pm37.73}$&$44.31_{\pm33.84}$&$44.6_{\pm34.06}$&$38.91_{\pm29.72}$&$34.25_{\pm26.16}$&$27.57_{\pm21.06}$&$37.43_{\pm28.59}$\\
&TGCN&$71.56_{\pm54.65}$&$49.2_{\pm37.58}$&$120.31_{\pm91.89}$&$48.37_{\pm36.94}$&$36.76_{\pm28.08}$&$32.09_{\pm24.51}$&$39.02_{\pm29.8}$\\
&GCLSTM&$63.9_{\pm48.8}$&$31.28_{\pm23.89}$&$28.84_{\pm22.03}$&$39.16_{\pm29.91}$&$35.03_{\pm26.76}$&$32.18_{\pm24.58}$&$37.87_{\pm28.93}$\\
&GConvLSTM&$68.73_{\pm52.49}$&$45.39_{\pm34.66}$&$44.43_{\pm33.94}$&$38.47_{\pm29.38}$&$33.75_{\pm25.77}$&$28.62_{\pm21.86}$&$37.96_{\pm28.99}$\\
&DyGrEncoder&$64.22_{\pm49.05}$&$43.72_{\pm33.39}$&$44.73_{\pm34.16}$&$43.73_{\pm33.4}$&$35.73_{\pm27.29}$&$30.42_{\pm23.23}$&$37.41_{\pm28.57}$\\
\midrule
\multirow{7}{*}{\rotatebox{90}{RRMSE(\%)}}&EvolveGCNH&$68.88_{\pm52.61}$&$20.26_{\pm15.47}$&$33.11_{\pm25.29}$&$36.88_{\pm28.17}$&$35.03_{\pm26.76}$&$26.91_{\pm20.55}$&$68.99_{\pm52.69}$\\
&EvolveGCNO&$17.01_{\pm12.99}$&$18.47_{\pm14.11}$&$28.97_{\pm22.12}$&$34.88_{\pm26.64}$&$32.23_{\pm24.61}$&$20.17_{\pm15.41}$&$59.82_{\pm45.69}$\\
&GConvGRU&$426.16_{\pm325.49}$&$1072.11_{\pm818.84}$&$1215.75_{\pm928.55}$&$763.66_{\pm583.25}$&$485.64_{\pm370.91}$&$255.02_{\pm194.77}$&$427.8_{\pm326.74}$\\
&TGCN&$3708.55_{\pm2832.45}$&$1159.31_{\pm885.44}$&$156814.93_{\pm119769.38}$&$1119.75_{\pm855.22}$&$621.0_{\pm474.3}$&$290.69_{\pm222.02}$&$451.93_{\pm345.17}$\\
&GCLSTM&$3210.27_{\pm2451.88}$&$184.79_{\pm141.14}$&$164.56_{\pm125.68}$&$786.29_{\pm600.54}$&$553.67_{\pm422.87}$&$291.57_{\pm222.69}$&$475.27_{\pm362.99}$\\
&GConvLSTM&$4672.71_{\pm3568.84}$&$1217.46_{\pm929.85}$&$1206.73_{\pm921.66}$&$695.01_{\pm530.83}$&$527.58_{\pm402.95}$&$326.17_{\pm249.12}$&$505.82_{\pm386.33}$\\
&DyGrEncoder&$3526.47_{\pm2693.38}$&$1093.03_{\pm834.81}$&$1365.47_{\pm1042.9}$&$1110.66_{\pm848.28}$&$525.05_{\pm401.02}$&$279.39_{\pm213.39}$&$465.82_{\pm355.78}$\\
\bottomrule
\end{tabular}
}
    \label{tab:graph_break}
\end{table}
\begin{table}[ht!]
    \centering
    \caption{Performance comparisons on low-frequency skill demand series with GNN-based methods.}
\resizebox{\textwidth}{!}{
\begin{tabular}{l|l|lllllll}
\toprule
&{Model}&Market&Region&L1-O&L2-O&R\&L1-O&R\&L2-O&Company\\
\midrule
\multirow{7}{*}{\rotatebox{90}{MAE}}&EvolveGCNH&$35.53_{\pm27.13}$&$2.28_{\pm1.74}$&$1.47_{\pm1.13}$&$0.55_{\pm0.42}$&$0.31_{\pm0.23}$&$0.14_{\pm0.11}$&$0.19_{\pm0.15}$\\
&EvolveGCNO&$27.87_{\pm21.29}$&$3.35_{\pm2.56}$&$1.69_{\pm1.29}$&$0.57_{\pm0.43}$&$0.31_{\pm0.24}$&$0.27_{\pm0.21}$&$0.17_{\pm0.13}$\\
&GConvGRU&$63.05_{\pm48.16}$&$3.06_{\pm2.34}$&$0.25_{\pm0.19}$&$0.18_{\pm0.14}$&$0.11_{\pm0.09}$&$0.05_{\pm0.04}$&$0.11_{\pm0.09}$\\
&TGCN&$7.37_{\pm5.63}$&$2.16_{\pm1.65}$&$0.13_{\pm0.1}$&$0.61_{\pm0.46}$&$0.26_{\pm0.2}$&$0.15_{\pm0.12}$&$0.15_{\pm0.11}$\\
&GCLSTM&$0.62_{\pm0.47}$&$0.41_{\pm0.32}$&$1.37_{\pm1.04}$&$0.15_{\pm0.11}$&$0.12_{\pm0.09}$&$0.09_{\pm0.07}$&$0.12_{\pm0.09}$\\
&GConvLSTM&$0.79_{\pm0.6}$&$0.43_{\pm0.33}$&$0.2_{\pm0.15}$&$0.13_{\pm0.1}$&$0.1_{\pm0.08}$&$0.06_{\pm0.05}$&$0.13_{\pm0.1}$\\
&DyGrEncoder&$1.04_{\pm0.79}$&$0.47_{\pm0.36}$&$0.24_{\pm0.18}$&$0.13_{\pm0.1}$&$0.15_{\pm0.11}$&$0.09_{\pm0.07}$&$0.14_{\pm0.11}$\\
\midrule
\multirow{7}{*}{\rotatebox{90}{RMSE}}&EvolveGCNH&$144.53_{\pm110.39}$&$11.08_{\pm8.46}$&$11.65_{\pm8.9}$&$5.52_{\pm4.22}$&$3.48_{\pm2.66}$&$1.7_{\pm1.3}$&$1.81_{\pm1.38}$\\
&EvolveGCNO&$83.5_{\pm63.77}$&$10.36_{\pm7.91}$&$12.82_{\pm9.79}$&$5.74_{\pm4.38}$&$3.8_{\pm2.9}$&$1.95_{\pm1.49}$&$1.65_{\pm1.26}$\\
&GConvGRU&$63.06_{\pm48.16}$&$3.65_{\pm2.79}$&$1.19_{\pm0.91}$&$0.87_{\pm0.66}$&$1.59_{\pm1.22}$&$0.81_{\pm0.62}$&$1.72_{\pm1.31}$\\
&TGCN&$21.13_{\pm16.14}$&$8.89_{\pm6.79}$&$1.13_{\pm0.86}$&$3.12_{\pm2.38}$&$2.32_{\pm1.77}$&$1.11_{\pm0.85}$&$1.78_{\pm1.36}$\\
&GCLSTM&$1.63_{\pm1.24}$&$2.48_{\pm1.89}$&$2.8_{\pm2.14}$&$0.86_{\pm0.66}$&$1.61_{\pm1.23}$&$0.83_{\pm0.63}$&$1.73_{\pm1.32}$\\
&GConvLSTM&$1.23_{\pm0.94}$&$2.1_{\pm1.6}$&$1.14_{\pm0.87}$&$0.85_{\pm0.65}$&$1.6_{\pm1.22}$&$0.81_{\pm0.62}$&$1.74_{\pm1.33}$\\
&DyGrEncoder&$1.49_{\pm1.14}$&$2.12_{\pm1.62}$&$1.15_{\pm0.88}$&$0.86_{\pm0.66}$&$1.59_{\pm1.22}$&$0.81_{\pm0.62}$&$1.73_{\pm1.32}$\\
\midrule
\multirow{7}{*}{\rotatebox{90}{SMAPE(\%)}}&EvolveGCNH&$47.39_{\pm36.19}$&$42.25_{\pm32.27}$&$34.5_{\pm26.35}$&$25.49_{\pm19.47}$&$20.27_{\pm15.48}$&$12.66_{\pm9.67}$&$21.3_{\pm16.27}$\\
&EvolveGCNO&$115.04_{\pm87.86}$&$89.67_{\pm68.48}$&$48.24_{\pm36.84}$&$28.83_{\pm22.02}$&$20.7_{\pm15.81}$&$29.63_{\pm22.63}$&$19.09_{\pm14.58}$\\
&GConvGRU&$131.13_{\pm100.15}$&$105.97_{\pm80.93}$&$26.95_{\pm20.58}$&$23.04_{\pm17.6}$&$13.04_{\pm9.96}$&$6.02_{\pm4.6}$&$11.92_{\pm9.1}$\\
&TGCN&$83.42_{\pm63.71}$&$48.87_{\pm37.32}$&$11.74_{\pm8.97}$&$28.63_{\pm21.87}$&$14.45_{\pm11.04}$&$16.13_{\pm12.32}$&$15.27_{\pm11.67}$\\
&GCLSTM&$42.8_{\pm32.69}$&$36.75_{\pm28.07}$&$43.27_{\pm33.05}$&$17.57_{\pm13.42}$&$14.03_{\pm10.72}$&$12.17_{\pm9.3}$&$13.44_{\pm10.26}$\\
&GConvLSTM&$53.85_{\pm41.13}$&$36.93_{\pm28.2}$&$21.67_{\pm16.55}$&$14.71_{\pm11.24}$&$10.07_{\pm7.69}$&$7.15_{\pm5.46}$&$14.91_{\pm11.39}$\\
&DyGrEncoder&$62.53_{\pm47.76}$&$39.84_{\pm30.43}$&$26.26_{\pm20.06}$&$15.61_{\pm11.92}$&$17.33_{\pm13.23}$&$12.71_{\pm9.7}$&$16.23_{\pm12.39}$\\
\midrule
\multirow{7}{*}{\rotatebox{90}{RRMSE(\%)}}&EvolveGCNH&$66.67_{\pm50.92}$&$67.56_{\pm51.6}$&$66.87_{\pm51.08}$&$67.29_{\pm51.39}$&$73.11_{\pm55.84}$&$73.29_{\pm55.97}$&$138.14_{\pm105.51}$\\
&EvolveGCNO&$66.67_{\pm50.92}$&$67.59_{\pm51.63}$&$66.81_{\pm51.03}$&$67.22_{\pm51.34}$&$69.95_{\pm53.43}$&$71.57_{\pm54.66}$&$119.39_{\pm91.18}$\\
&GConvGRU&$66.46_{\pm50.76}$&$75.73_{\pm57.84}$&$122.73_{\pm93.74}$&$153.14_{\pm116.96}$&$297.91_{\pm227.54}$&$242.77_{\pm185.42}$&$241.43_{\pm184.39}$\\
&TGCN&$66.51_{\pm50.8}$&$68.02_{\pm51.95}$&$676.51_{\pm516.7}$&$68.73_{\pm52.5}$&$89.88_{\pm68.65}$&$92.87_{\pm70.93}$&$190.19_{\pm145.26}$\\
&GCLSTM&$60.79_{\pm46.43}$&$103.27_{\pm78.88}$&$70.15_{\pm53.58}$&$169.24_{\pm129.26}$&$359.25_{\pm274.38}$&$214.34_{\pm163.7}$&$272.3_{\pm207.97}$\\
&GConvLSTM&$65.93_{\pm50.36}$&$189.03_{\pm144.38}$&$156.39_{\pm119.44}$&$166.51_{\pm127.18}$&$345.76_{\pm264.08}$&$287.94_{\pm219.92}$&$285.4_{\pm217.98}$\\
&DyGrEncoder&$60.35_{\pm46.09}$&$171.99_{\pm131.36}$&$167.79_{\pm128.15}$&$171.6_{\pm131.06}$&$325.47_{\pm248.58}$&$279.07_{\pm213.14}$&$269.12_{\pm205.54}$\\
\bottomrule
\end{tabular}
}
    \label{tab:graph_low}
\end{table}

\begin{table}[ht!]
    \centering
    \caption{Performance comparisons on skill demand proportion forecasting.}
\resizebox{\textwidth}{!}{
\begin{tabular}{l|llllllll}
\toprule
&Model&Market&Region&L1-O&L2-O&R\&L1-O&R\&L2-O&Company\\
\midrule
\multirow{16}{*}{\rotatebox{90}{MAE(\%)}}&LSTM&$0.14\pm_{0.0}$&$0.49\pm_{0.01}$&$1.21\pm_{0.03}$&$2.26\pm_{0.05}$&$2.43\pm_{0.03}$&$3.51\pm_{0.06}$&$2.69\pm_{0.03}$\\
&SegRNN&$0.15\pm_{0.02}$&$1.25\pm_{0.02}$&$2.63\pm_{0.11}$&$4.2\pm_{0.04}$&$6.6\pm_{1.12}$&$10.16\pm_{2.44}$&$5.52\pm_{0.53}$\\
&CHGH&$0.1\pm_{0.01}$&$0.19\pm_{0.0}$&$0.35\pm_{0.0}$&$0.64\pm_{0.01}$&$0.7\pm_{0.0}$&$0.91\pm_{0.07}$&$1.03\pm_{0.01}$\\
&PreDyGAE&$0.08\pm_{0.01}$&$0.09\pm_{0.0}$&$0.12\pm_{0.02}$&$0.23\pm_{0.02}$&$0.18\pm_{0.0}$&$0.27\pm_{0.15}$&$0.29\pm_{0.16}$\\
&Transformer&$0.62\pm_{0.02}$&$3.89\pm_{0.02}$&$10.87\pm_{0.05}$&$20.58\pm_{0.01}$&$22.02\pm_{0.01}$&$31.26\pm_{0.02}$&$23.47\pm_{0.01}$\\
&Autoformer&$1.29\pm_{0.14}$&$8.84\pm_{0.15}$&$24.5\pm_{0.22}$&$43.99\pm_{0.94}$&$47.53\pm_{0.67}$&$66.32\pm_{1.02}$&$50.02\pm_{0.66}$\\
&Informer&$0.55\pm_{0.02}$&$3.83\pm_{0.03}$&$10.86\pm_{0.03}$&$20.6\pm_{0.01}$&$22.04\pm_{0.01}$&$31.23\pm_{0.01}$&$23.48\pm_{0.0}$\\
&Reformer&$0.56\pm_{0.01}$&$3.85\pm_{0.01}$&$10.9\pm_{0.03}$&$20.65\pm_{0.02}$&$22.01\pm_{0.01}$&$31.25\pm_{0.01}$&$23.48\pm_{0.0}$\\
&FEDformer&$1.42\pm_{0.04}$&$8.58\pm_{0.18}$&$25.11\pm_{0.59}$&$44.15\pm_{0.6}$&$47.07\pm_{0.66}$&$65.95\pm_{0.23}$&$50.0\pm_{0.07}$\\
&NStransformer&$0.06\pm_{0.0}$&$0.09\pm_{0.0}$&$0.11\pm_{0.0}$&$0.19\pm_{0.0}$&$0.24\pm_{0.0}$&$0.34\pm_{0.0}$&$0.34\pm_{0.0}$\\
&PatchTST&$0.06\pm_{0.0}$&$0.09\pm_{0.01}$&$0.11\pm_{0.02}$&$0.18\pm_{0.04}$&$0.23\pm_{0.05}$&$0.33\pm_{0.07}$&$0.33\pm_{0.06}$\\
&DLinear&$0.27\pm_{0.15}$&$1.65\pm_{1.07}$&$4.63\pm_{3.09}$&$8.76\pm_{5.86}$&$9.4\pm_{6.26}$&$13.33\pm_{8.89}$&$10.05\pm_{6.65}$\\
&TSMixer&$0.85\pm_{0.03}$&$3.15\pm_{0.69}$&$7.57\pm_{1.43}$&$18.88\pm_{0.49}$&$15.12\pm_{3.85}$&$18.76\pm_{2.85}$&$17.77\pm_{1.58}$\\
&FreTS&$0.12\pm_{0.0}$&$0.31\pm_{0.04}$&$0.69\pm_{0.11}$&$1.26\pm_{0.21}$&$1.37\pm_{0.22}$&$1.94\pm_{0.32}$&$1.54\pm_{0.24}$\\
&FiLM&$0.06\pm_{0.01}$&$0.09\pm_{0.01}$&$0.12\pm_{0.01}$&$0.19\pm_{0.02}$&$0.25\pm_{0.03}$&$0.35\pm_{0.05}$&$0.34\pm_{0.05}$\\
&Koopa&$0.06\pm_{0.0}$&$0.08\pm_{0.0}$&$0.09\pm_{0.0}$&$0.15\pm_{0.01}$&$0.2\pm_{0.01}$&$0.27\pm_{0.02}$&$0.28\pm_{0.01}$\\
\midrule
\multirow{16}{*}{\rotatebox{90}{RMSE(\%)}}&LSTM&$0.61\pm_{0.01}$&$1.74\pm_{0.04}$&$2.89\pm_{0.08}$&$4.06\pm_{0.09}$&$4.16\pm_{0.05}$&$5.21\pm_{0.1}$&$4.43\pm_{0.05}$\\
&SegRNN&$1.32\pm_{0.17}$&$4.98\pm_{0.11}$&$6.49\pm_{0.22}$&$7.7\pm_{0.09}$&$11.23\pm_{1.78}$&$14.58\pm_{3.23}$&$9.18\pm_{0.83}$\\
&CHGH&$0.36\pm_{0.01}$&$0.49\pm_{0.0}$&$0.7\pm_{0.0}$&$1.12\pm_{0.02}$&$1.14\pm_{0.0}$&$2.11\pm_{0.32}$&$2.31\pm_{0.07}$\\
&PreDyGAE&$0.23\pm_{0.0}$&$0.29\pm_{0.0}$&$0.36\pm_{0.0}$&$0.85\pm_{0.0}$&$0.91\pm_{0.0}$&$1.15\pm_{1.05}$&$1.29\pm_{0.72}$\\
&Transformer&$6.22\pm_{0.12}$&$16.65\pm_{0.08}$&$27.9\pm_{0.14}$&$38.45\pm_{0.03}$&$39.77\pm_{0.01}$&$47.4\pm_{0.02}$&$40.97\pm_{0.01}$\\
&Autoformer&$15.11\pm_{2.51}$&$38.94\pm_{0.61}$&$63.72\pm_{0.49}$&$82.56\pm_{1.83}$&$86.21\pm_{1.22}$&$100.99\pm_{1.58}$&$87.83\pm_{1.13}$\\
&Informer&$5.63\pm_{0.15}$&$16.41\pm_{0.13}$&$27.89\pm_{0.05}$&$38.5\pm_{0.03}$&$39.79\pm_{0.01}$&$47.36\pm_{0.01}$&$40.98\pm_{0.0}$\\
&Reformer&$5.62\pm_{0.05}$&$16.45\pm_{0.05}$&$27.96\pm_{0.11}$&$38.58\pm_{0.02}$&$39.76\pm_{0.03}$&$47.38\pm_{0.02}$&$40.98\pm_{0.02}$\\
&FEDformer&$17.24\pm_{0.68}$&$37.52\pm_{0.77}$&$64.93\pm_{1.42}$&$82.85\pm_{1.16}$&$85.38\pm_{1.21}$&$100.35\pm_{0.34}$&$87.74\pm_{0.1}$\\
&NStransformer&$0.21\pm_{0.0}$&$0.28\pm_{0.0}$&$0.35\pm_{0.0}$&$0.77\pm_{0.01}$&$0.86\pm_{0.0}$&$1.69\pm_{0.0}$&$1.24\pm_{0.0}$\\
&PatchTST&$0.21\pm_{0.01}$&$0.28\pm_{0.02}$&$0.35\pm_{0.03}$&$0.75\pm_{0.04}$&$0.84\pm_{0.06}$&$1.65\pm_{0.1}$&$1.22\pm_{0.07}$\\
&DLinear&$2.21\pm_{1.49}$&$6.05\pm_{4.1}$&$10.27\pm_{6.97}$&$14.16\pm_{9.59}$&$14.64\pm_{9.91}$&$17.51\pm_{11.77}$&$15.11\pm_{10.19}$\\
&TSMixer&$10.5\pm_{0.04}$&$14.32\pm_{2.43}$&$21.09\pm_{3.64}$&$34.92\pm_{1.37}$&$25.27\pm_{5.66}$&$26.48\pm_{3.26}$&$31.07\pm_{2.82}$\\
&FreTS&$0.47\pm_{0.02}$&$0.96\pm_{0.07}$&$1.56\pm_{0.12}$&$2.22\pm_{0.17}$&$2.29\pm_{0.17}$&$2.99\pm_{0.2}$&$2.54\pm_{0.17}$\\
&FiLM&$0.21\pm_{0.0}$&$0.28\pm_{0.0}$&$0.35\pm_{0.02}$&$0.76\pm_{0.04}$&$0.85\pm_{0.06}$&$1.67\pm_{0.11}$&$1.22\pm_{0.05}$\\
&Koopa&$0.21\pm_{0.0}$&$0.27\pm_{0.0}$&$0.33\pm_{0.01}$&$0.72\pm_{0.01}$&$0.79\pm_{0.02}$&$1.57\pm_{0.04}$&$1.17\pm_{0.01}$\\
\midrule
\multirow{16}{*}{\rotatebox{90}{SMAPE(\%)}}&LSTM&$0.26\pm_{0.0}$&$0.92\pm_{0.02}$&$2.25\pm_{0.05}$&$4.21\pm_{0.09}$&$4.52\pm_{0.05}$&$6.5\pm_{0.11}$&$5.0\pm_{0.05}$\\
&SegRNN&$0.27\pm_{0.03}$&$2.09\pm_{0.03}$&$4.55\pm_{0.17}$&$7.38\pm_{0.06}$&$11.1\pm_{1.63}$&$16.79\pm_{3.54}$&$9.58\pm_{0.83}$\\
&CHGH&$0.19\pm_{0.01}$&$0.36\pm_{0.0}$&$0.69\pm_{0.0}$&$1.24\pm_{0.02}$&$1.36\pm_{0.0}$&$1.78\pm_{0.01}$&$1.77\pm_{0.06}$\\
&PreDyGAE&$0.09\pm_{0.0}$&$0.19\pm_{0.0}$&$0.22\pm_{0.0}$&$0.28\pm_{0.0}$&$0.39\pm_{0.0}$&$0.64\pm_{0.27}$&$0.57\pm_{0.3}$\\
&Transformer&$0.81\pm_{0.02}$&$4.78\pm_{0.01}$&$13.28\pm_{0.04}$&$25.1\pm_{0.0}$&$26.89\pm_{0.01}$&$38.12\pm_{0.01}$&$28.71\pm_{0.01}$\\
&Autoformer&$1.13\pm_{0.02}$&$7.42\pm_{0.05}$&$20.74\pm_{0.1}$&$38.35\pm_{0.38}$&$41.28\pm_{0.29}$&$58.08\pm_{0.43}$&$43.84\pm_{0.29}$\\
&Informer&$0.75\pm_{0.02}$&$4.74\pm_{0.02}$&$13.27\pm_{0.03}$&$25.12\pm_{0.0}$&$26.9\pm_{0.01}$&$38.1\pm_{0.01}$&$28.72\pm_{0.0}$\\
&Reformer&$0.75\pm_{0.01}$&$4.75\pm_{0.01}$&$13.31\pm_{0.02}$&$25.16\pm_{0.02}$&$26.87\pm_{0.01}$&$38.12\pm_{0.01}$&$28.72\pm_{0.0}$\\
&FEDformer&$1.17\pm_{0.01}$&$7.35\pm_{0.08}$&$21.02\pm_{0.24}$&$38.42\pm_{0.24}$&$41.09\pm_{0.28}$&$57.94\pm_{0.1}$&$43.84\pm_{0.03}$\\
&NStransformer&$0.11\pm_{0.0}$&$0.17\pm_{0.0}$&$0.22\pm_{0.0}$&$0.35\pm_{0.0}$&$0.45\pm_{0.0}$&$0.61\pm_{0.0}$&$0.63\pm_{0.0}$\\
&PatchTST&$0.11\pm_{0.01}$&$0.17\pm_{0.02}$&$0.21\pm_{0.05}$&$0.34\pm_{0.09}$&$0.44\pm_{0.1}$&$0.59\pm_{0.14}$&$0.61\pm_{0.11}$\\
&DLinear&$0.43\pm_{0.19}$&$2.58\pm_{1.39}$&$7.19\pm_{4.0}$&$13.59\pm_{7.59}$&$14.59\pm_{8.11}$&$20.67\pm_{11.52}$&$15.62\pm_{8.62}$\\
&TSMixer&$0.92\pm_{0.03}$&$4.05\pm_{0.67}$&$10.03\pm_{1.36}$&$23.97\pm_{1.21}$&$21.46\pm_{4.26}$&$27.6\pm_{3.5}$&$23.83\pm_{1.45}$\\
&FreTS&$0.22\pm_{0.01}$&$0.59\pm_{0.07}$&$1.31\pm_{0.21}$&$2.4\pm_{0.41}$&$2.63\pm_{0.43}$&$3.69\pm_{0.62}$&$2.95\pm_{0.46}$\\
&FiLM&$0.11\pm_{0.02}$&$0.17\pm_{0.02}$&$0.22\pm_{0.02}$&$0.36\pm_{0.03}$&$0.46\pm_{0.06}$&$0.63\pm_{0.09}$&$0.64\pm_{0.09}$\\
&Koopa&$0.1\pm_{0.0}$&$0.15\pm_{0.01}$&$0.18\pm_{0.01}$&$0.28\pm_{0.01}$&$0.37\pm_{0.02}$&$0.49\pm_{0.03}$&$0.52\pm_{0.03}$\\
\midrule
\multirow{16}{*}{\rotatebox{90}{RRMSE(\%)}}&LSTM&$43.76\pm_{0.35}$&$81.48\pm_{0.6}$&$85.53\pm_{0.64}$&$91.34\pm_{0.35}$&$93.23\pm_{0.15}$&$97.48\pm_{0.09}$&$96.6\pm_{0.07}$\\
&SegRNN&$72.41\pm_{4.5}$&$96.92\pm_{0.17}$&$96.4\pm_{0.21}$&$96.99\pm_{0.11}$&$98.79\pm_{0.46}$&$99.24\pm_{0.25}$&$98.15\pm_{0.28}$\\
&CHGH&$27.4\pm_{0.7}$&$37.66\pm_{0.07}$&$37.45\pm_{0.03}$&$53.25\pm_{0.58}$&$58.32\pm_{0.03}$&$80.91\pm_{5.07}$&$70.89\pm_{2.0}$\\
&PreDyGAE&$20.64\pm_{0.0}$&$28.83\pm_{0.0}$&$40.02\pm_{0.01}$&$40.12\pm_{0.0}$&$47.5\pm_{0.0}$&$76.9\pm_{42.81}$&$66.68\pm_{33.49}$\\
&Transformer&$97.99\pm_{0.07}$&$99.73\pm_{0.01}$&$99.8\pm_{0.0}$&$99.89\pm_{0.0}$&$99.92\pm_{0.0}$&$99.97\pm_{0.0}$&$99.96\pm_{0.0}$\\
&Autoformer&$99.63\pm_{0.12}$&$99.95\pm_{0.0}$&$99.96\pm_{0.0}$&$99.97\pm_{0.0}$&$99.98\pm_{0.0}$&$99.99\pm_{0.0}$&$99.98\pm_{0.0}$\\
&Informer&$97.59\pm_{0.12}$&$99.72\pm_{0.0}$&$99.8\pm_{0.0}$&$99.89\pm_{0.0}$&$99.92\pm_{0.0}$&$99.97\pm_{0.0}$&$99.96\pm_{0.0}$\\
&Reformer&$97.57\pm_{0.03}$&$99.72\pm_{0.01}$&$99.81\pm_{0.0}$&$99.89\pm_{0.0}$&$99.92\pm_{0.0}$&$99.97\pm_{0.0}$&$99.96\pm_{0.0}$\\
&FEDformer&$99.75\pm_{0.02}$&$99.95\pm_{0.0}$&$99.96\pm_{0.0}$&$99.97\pm_{0.0}$&$99.98\pm_{0.0}$&$99.99\pm_{0.0}$&$99.98\pm_{0.0}$\\
&NStransformer&$16.44\pm_{0.44}$&$22.17\pm_{0.31}$&$19.66\pm_{0.16}$&$37.34\pm_{0.26}$&$45.81\pm_{0.26}$&$75.98\pm_{0.13}$&$61.84\pm_{0.13}$\\
&PatchTST&$16.32\pm_{0.88}$&$21.85\pm_{1.31}$&$19.5\pm_{1.44}$&$36.66\pm_{1.59}$&$44.97\pm_{1.95}$&$75.59\pm_{0.49}$&$61.55\pm_{1.34}$\\
&DLinear&$78.33\pm_{15.19}$&$95.41\pm_{3.54}$&$96.69\pm_{2.57}$&$97.94\pm_{1.58}$&$98.55\pm_{1.11}$&$99.37\pm_{0.44}$&$98.87\pm_{0.84}$\\
&TSMixer&$99.29\pm_{0.02}$&$99.61\pm_{0.13}$&$99.64\pm_{0.11}$&$99.86\pm_{0.02}$&$99.8\pm_{0.11}$&$99.85\pm_{0.0}$&$99.91\pm_{0.05}$\\
&FreTS&$34.81\pm_{1.43}$&$61.32\pm_{3.03}$&$66.35\pm_{3.08}$&$77.25\pm_{2.53}$&$81.41\pm_{2.37}$&$92.32\pm_{1.5}$&$90.05\pm_{1.64}$\\
&FiLM&$16.49\pm_{0.39}$&$21.81\pm_{0.76}$&$19.48\pm_{0.05}$&$36.44\pm_{0.97}$&$44.79\pm_{2.15}$&$75.91\pm_{10.46}$&$61.27\pm_{8.05}$\\
&Koopa&$16.47\pm_{0.04}$&$21.47\pm_{0.18}$&$18.71\pm_{0.07}$&$35.58\pm_{0.07}$&$43.65\pm_{0.38}$&$75.85\pm_{2.58}$&$61.26\pm_{1.98}$\\
\bottomrule
\end{tabular}
}
    \label{tab:Propotion}
\end{table}

\subsection{Skill Co-occurrence Graph Enhanced Job Skill Demand Forecasting}

% \paragraph{Skill Co-occurrence Graph.}
In the task of job skill demand forecasting, fully leveraging the inter-relationships among different skills is beneficial for downstream tasks. Therefore, we construct a prior graph with co-occurrence frequency from the training data to include as a dataset component. 
Given a set of granularities $i,j,\ldots,k$, we constructed the skill co-occurrence graph as $\mathcal{G}^{i,j,\ldots,k} = (\mathcal{V}^{i,j,\ldots,k}, \mathcal{E}^{i,j,\ldots,k})$, where $\mathcal{V}^{i,j,\ldots,k}$ is the extended skill set under the multiple granularities. The edge weight $e_{v, v^{\prime}}\in \mathcal{E}^{i,j,...,k}$ between nodes $v$ and $v^{\prime}$ is determined by the co-occurrence frequency of the node pair $v, v^{\prime}$ in the job advertisement data for training $\mathcal{P}^{train}$. Specifically, given $v=(a^i, a^j, \ldots, a^k, s)$ and $v^{\prime} = (a^{i\prime}, a^{j\prime}, ..., a^{k\prime}, s^{\prime})$, $e_{v, v^{\prime}}$ is calculated as:
\begin{equation}
e_{v, v^{\prime}} = \sum_{p\in \mathcal{P}^{train}} \prod_{x \in \{a^i, a^j, \ldots, a^k, a^{i\prime}, a^{j\prime}, \ldots, a^{k\prime}, s, s^{\prime}\}} \mathbf{1}_p(x\in p). 
\end{equation}

This information will serve as prior knowledge, reflecting global inter-skill dependency patterns.

\paragraph{Benchmark Models}
To fully utilize the prior information from the co-occurency graph, we introduce several GNN-based methods for multivariate time series prediction. These methods leverage GNNs to extract the influences between different variables, effectively capturing the relationships among various time series. The specific models are as follows:

\begin{itemize}
\item \textbf{EvolveGCN} \citep{EvolveGCN}: EvolveGCN introduces a recurrent mechanism to update the network parameters, as GCN parameters, for capturing the dynamism of the graphs. Two methods are introduced: EvolveGCNH, which learns the weight matrix of the graph at each time step as a hidden state, and EvolveGCNO, which directly employs the weight evolution as a hidden state output, decoupled from node embedding.

\item \textbf{GConvGRU} \citep{GconvLSTM}: This model integrates convolutional neural networks (CNNs) on graphs to identify spatial structures and recurrent neural networks (RNNs) to detect dynamic patterns. Two architectures, GConvGRU and GConvLSTM, are explored for the Graph Convolutional Recurrent Network (GCRN).

\item \textbf{TGCN} \citep{zhao_t-gcn_2020}: The temporal graph convolutional network (T-GCN) model, which is in combination with the graph convolutional network (GCN) and gated recurrent unit (GRU). Specifically, the GCN is used to learn complex topological structures to capture spatial dependence and the gated recurrent unit is used to learn dynamic changes of traffic data to capture temporal dependence.

\item \textbf{GCLSTM} \citep{GC-LSTM}: GCLSTM is an end-to-end model integrating a Graph Convolution Network (GCN) embedded Long Short-Term Memory network (LSTM) for dynamic network link prediction. The GCN captures local structural properties, while the LSTM learns temporal features across snapshots of a dynamic network.

\item \textbf{DyGrEncoder} \citep{dygrencoder}: This approach combines a sequence-to-sequence encoder-decoder model with gated graph neural networks (GGNNs) and long short-term memory networks (LSTMs). The encoder captures temporal dynamics in an evolving graph, and the decoder reconstructs the dynamics using the encoded representation.

\end{itemize}
We implement these benchmark models using the PyG library~\footnote{https://github.com/benedekrozemberczki/pytorch\_geometric\_temporal} and demonstrate the effectiveness of these GNN-based methods in skill demand forecasting.

\paragraph{Results}
We have implemented a series of graph-based multivariate time series forecasting methods based on the co-occurrence graph and verified their experimental effects under the three scenarios discussed above. Firstly, Table \ref{tab:graph_overall} presents the overall performance of the methods based on the co-occurrence graph for skill demand forecasting. It is observed that the prediction accuracy significantly declines across the overall labor market. However, as the granularity of the forecast becomes finer, the model performance improves, and at a finer granularity, the EvolveGCN method outperforms the state-of-the-art (SOTA) methods mentioned in the main text considerably. We analyze that the finer the granularity, the more accurately the co-occurrence graph reflects the associations between skills, while coarser granularity might introduce excessive noise leading to decreased model performance. The fine-grained co-occurrence graph accurately reflects the interrelationships between skills at different granularities, which aids in enhancing the model's prediction accuracy. Secondly, we find significant differences in the RRMSE metric among these methods, with EvolveGCN showing superior performance because it can learn the evolution of GCN parameter weights over time, thus capturing the evolving dependencies among edges. Therefore, based on the provided co-occurrence graph, it can effectively learn the evolution of skill relationships, which is beneficial for dynamic prediction of skill demand.

For skill demand forecasting in scenarios involving structural breaks, as shown in Table \ref{tab:graph_break}, the improvements in the methods based on the co-occurrence graph are greater than those in the overall skill demand forecasting task. This suggests that skills experiencing structural breaks have strong interconnections, and the co-occurrence graph helps the model to identify the patterns of skill demand sequences that are likely to undergo structural breaks, thus further enhancing the prediction effectiveness for this category of skills.

In the task of predicting low-frequency skills, as shown in Table \ref{tab:graph_low}, methods like GConvLSTM significantly outperform EvolveGCN. This is due to the sparse observable data for these skills, which leads to sparse connectivity edges on the co-occurrence graph. 

\subsection{Job Skill Demand Proportion Forecasting}
In the main text, we discussed the issue of skill demand prediction. However, consider a scenario where the number of skill postings for a particular occupation is very low, leading to a low demand for that occupational skills. Nevertheless, these skills might constitute a significant portion of the profession's core competencies. Therefore, using skill demand alone may not adequately measure the importance of these skills within the occupation. To address this, we introduce an extended dataset that includes the skill demand propotion. We define the skill demand proportion as:
\begin{equation}
R_{s, t}^i = [R_{s, t, a^i}^i]_{a^i \in \mathcal{A}^i}, \ \ R_{s, t, a^i}^i =\frac{\sum_{p \in \mathcal{P}_t} \mathbf{1}(s \in p) \cdot \mathbf{1}(a^i \in p)}{\sum_{p \in \mathcal{P}_t} \mathbf{1}(a^i \in p)},
\end{equation}
where $a^i \in p$ represents a job advertisement $p$ containing the attribute $a^i$ under granularity $i$. Similarly, we can further define skill demand proportions $R_{s, t}^{i,j,...,k}$ across multiple granularities $\{i, j, ..., k\}$ by calculating:
\begin{equation}
\label{eq::proportion}
R_{s, t, \overline{a}}^{i,j,...,k} = \frac{\sum_{p \in \mathcal{P}_t} \mathbf{1}(s \in p) \cdot \mathbf{1}(a^i \in p \land a^j \in p \land ... \land a^k \in p)}{\sum_{p \in \mathcal{P}_t} \mathbf{1}(a^i \in p \land a^j \in p \land ... \land a^k \in p)} ,
\end{equation}
where $\overline{a} = \{a^i, a^j, ..., a^k\}$, $a^i \in \mathcal{A}^i, a^j \in \mathcal{A}^j, ..., a^k \in \mathcal{A}^k$, and $R^{i,j,\ldots,k}_{s,t} \in \mathbb{R}^{ |\mathcal{A}^i||\mathcal{A}^j|\ldots|\mathcal{A}^k|}$.

\paragraph{Results}

We continue to utilize the benchmark models described in the main text for this task, and the results, as shown in Table \ref{tab:Propotion}, lead to the following conclusions:
Firstly, the best-performing model on the task of forecasting the proportion of skill demand is Koopa. This model, integrating time series decomposition and Fourier transformations, effectively captures the distribution changes in demand proportions. Secondly, there is a significant variation in performance across models in this task. For example, models like DLinear perform poorly on this task, though they are reasonably effective in skill demand forecasting. We analyze that predicting percentages is distinct from forecasting skill demand, as percentage predictions are also influenced by the demand for other skills at the same granularity. Therefore, simple linear models are not advantageous for capturing the complex interrelations and influences among multiple pieces of information.

\section{Data Structure and Components}
\label{data::intro}
Our dataset comprises five components for each granularity level: job skill demand sequences, job skill demand proportion sequences, ID mapping index, the indexes of skills with structural breaks, and skill co-occurrence graph. Each component is structured as follows:
(1) \textit{Job Skill Demand Sequences:} These are presented in tabular files, where each row represents a specific skill, and each column corresponds to a different time slice (month). Each cell within the table contains a numerical value that reflects the demand for the respective skill during that month.
(ii) \textit{Job Skill Demand Proportion Sequences:} This component is also formatted in tabular files similar to the skill demand sequences. However, each cell in these tables displays a value between 0 and 1, representing the proportion of demand defined in Eq \ref{eq::proportion}. This provides a normalized view of skill demand across different granularities.
(iii) \textit{ID mapping index:} In the dataset, various elements such as regions, occupations, companies, and skills are represented using unique identifiers (IDs) for the convenience of experimentation and analysis. An index table is provided that maps each ID to the actual names of regions, occupations, and skills, facilitating clear and effective data interpretation. The names of companies, however, are withheld due to potential privacy concerns. The remaining mapping tables are not publicly available on GitHub. Researchers requiring access to these mapping tables, excluding company-related data, may contact the first author via email to submit a request.
(iv) \textit{Indexes of skills with structural breaks:} In the provided dataset, data concerning skills that have experienced structural breaks are organized in JSON format. Each granularity level is represented by a separate JSON file, which contains a list of indexes. These indexes correspond to the skills that have undergone structural breaks and can be directly mapped to the skill indexes in the skill demand sequences. The purpose of supplying this data is to facilitate research on the demand trends of skills that have exhibited structural breaks, enabling a detailed analysis of their demand dynamics over time.
(v) \textit{Skill Co-occurrence Graph:} This data is provided as a set of triples (skill ID\_1, skill ID\_2, frequency of co-occurrence), forming a collection that outlines the co-occurrence relationships between skills. Each triple indicates how frequently two skills are mentioned or required together within the job advertisements in the training data, serving as a prior knowledge graph to enhance predictive modeling by capturing relationships between skills.

% \section{Limitations}
% This work presents several areas for future exploration and challenges that can be further addressed:
% Firstly, this paper presents the co-occurrence graph as auxiliary input in the appendix and does not comprehensively discuss its impact on skill demand forecasting or the related model development in the main text. This aspect involves validating the positive significance of the co-occurrence graph for skill demand forecasting and exploring its impact on model performance, which requires comprehensive and detailed data analysis and model testing. Additionally, constructing static or dynamic co-occurrence graphs is a strategy worth considering and experimenting with. Overall, the use of co-occurrence graphs represents a promising area for future research and exploration.

% Secondly, this paper identifies the phenomenon of structural breaks within skill demand sequences but lacks a targeted benchmark model to address this specific forecasting challenge. How to detect skills that will undergo structural breaks and to predict them accurately remains an area worthy of exploration.

% \input{8.check_list}

\end{document}